\pdfoutput=1

\documentclass[11pt]{article}

\usepackage{ACL2023}

\usepackage{times}
\usepackage{latexsym}

\usepackage[T1]{fontenc}

\usepackage[utf8]{inputenc}

\usepackage{microtype}

\usepackage{inconsolata}

\usepackage{amsmath}
\usepackage{times}
\usepackage{framed}
\usepackage{ragged2e}
\usepackage{bm}
\usepackage{soul}
\usepackage{latexsym}
\usepackage{subfigure}
\usepackage{amsthm}
\usepackage{graphicx}
\usepackage{float}
\usepackage{longtable}
\usepackage{booktabs} 
\usepackage{multirow}
\usepackage{multicol}
\usepackage{amssymb}
\usepackage{dashbox}%
\usepackage{multirow}
\usepackage{textcomp}
\usepackage{tcolorbox}
\usepackage{bbding}
\usepackage[ruled, vlined, linesnumbered]{algorithm2e}
\usepackage{mathtools}
\usepackage{adjustbox}
\usepackage[ruled,vlined]{algorithm2e}
\usepackage{color, soul}
\usepackage{pifont}
\usepackage{array}
\newcolumntype{P}[1]{>{\centering\arraybackslash}p{#1}}
\newcolumntype{M}[1]{>{\centering\arraybackslash}m{#1}}
\usepackage{cleveref}
\crefname{section}{§}{§§}
\Crefname{section}{§}{§§}
\crefname{figure}{Figure}{Figure}
\Crefname{figure}{Figure}{Figure}
\crefname{table}{Table}{Table}
\Crefname{table}{Table}{Table}

\newcommand\acedygie{\texttt{ACE-DYGIE}\xspace}
\newcommand\aceoneie{\texttt{ACE-OneIE}\xspace}
\newcommand\acefull{\texttt{ACE-Full}\xspace}

\title{The Devil is in the Details: On the Pitfalls of Event Extraction Evaluation}

\author{Hao Peng$^{1}$\thanks{\quad Equal contribution. Random Order.}\hspace{0.5em}, Xiaozhi Wang$^{1*}$, 
Feng Yao$^{2*}$, Kaisheng Zeng$^{1}$, \\
\textbf{Lei Hou$^{1,3}$, Juanzi Li$^{1,3}$\thanks{\quad Corresponding author: J.Li}\hspace{0.5em}, Zhiyuan Liu$^{1,3}$, Weixing Shen$^{2}$} \\ 
$^1$Department of Computer Science and Technology, BNRist; \\
$^2$School of Law, Institute for AI and Law; \\
$^3$KIRC, Institute for Artificial Intelligence,\\
Tsinghua University, Beijing, 100084, China\\
{\tt \{peng-h21, wangxz20, yaof20\}@mails.tsinghua.edu.cn}
}

\begin{document}
\maketitle
\begin{abstract}
Event extraction (EE) is a crucial task aiming at extracting events from texts, which includes two subtasks: event detection (ED) and event argument extraction (EAE). In this paper, we check the reliability of EE evaluations and identify three major pitfalls: (1) The \textbf{data preprocessing discrepancy} makes the evaluation results on the same dataset not directly comparable, but the data preprocessing details are not widely noted and specified in papers. (2) The \textbf{output space discrepancy} of different model paradigms makes different-paradigm EE models lack grounds for comparison and also leads to unclear mapping issues between predictions and annotations. (3) The \textbf{absence of pipeline evaluation} of many EAE-only works makes them hard to be directly compared with EE works and may not well reflect the model performance in real-world pipeline scenarios. We demonstrate the significant influence of these pitfalls through comprehensive meta-analyses of recent papers and empirical experiments. To avoid these pitfalls, we suggest a series of remedies, including specifying data preprocessing, standardizing outputs, and providing pipeline evaluation results. To help implement these remedies, we develop a consistent evaluation framework \textsc{OmniEvent}, which can be obtained from \url{https://github.com/THU-KEG/OmniEvent}. 
\end{abstract}

\section{Introduction}
\begin{figure}[!t]
    \centering
    \includegraphics[width=\linewidth]{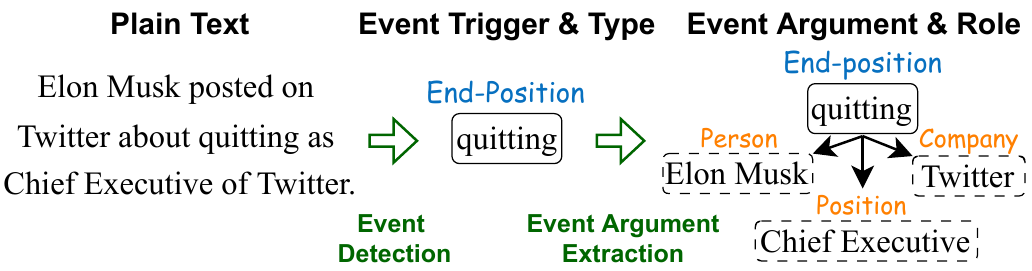}
    \caption{An illustration for the event extraction (EE) pipeline, including two stages: event detection (ED) and event argument extraction (EAE).}
    \label{fig:pipeline}
\end{figure}

Event extraction (EE) is a fundamental information extraction task aiming at extracting structural event knowledge from plain texts. As illustrated in \cref{fig:pipeline}, it is typically formalized as a two-stage pipeline~\citep{ahn2006stages}. The first subtask, event detection (ED), is to detect the event triggers (keywords or phrases evoking events, e.g., \textit{quitting} in \cref{fig:pipeline}) and classify their event types (e.g., \texttt{End-Position}). The second subtask, event argument extraction (EAE), is to extract corresponding event arguments and their roles (e.g., \textit{Elon Musk} and its argument role \texttt{Person}) based on the first-stage ED results. 

Since events play an important role in human language understanding and broad applications benefit from structural event knowledge~\citep{ji-grishman-2011-knowledge,glavavs2014event,hogenboom2016survey,zhang2020aser}, EE has attracted much research attention, and novel models have been continually developed. Beyond the conventional paradigms like classification~\citep{chen2015event,wang-etal-2021-cleve} and sequence labeling~\citep{nguyen-etal-2016-joint-event,chen-etal-2018-collective}, new model paradigms such as span prediction~\citep{liu-etal-2020-event,du-cardie-2020-event} and conditional generation~\citep{lu-etal-2021-text2event,li-etal-2021-document} are proposed. These sophisticated models push evaluation results to increasingly high levels.

However, due to the complex input/output formats and task pipeline of EE, there are some hidden pitfalls in EE evaluations, which are rarely noted and discussed in EE papers~\citep{wadden-etal-2019-entity,wang-etal-2020-maven,wang-etal-2022-query}. These pitfalls make many competing EE methods actually lack grounds for comparison, and the reported scores cannot reflect real-world model performances well.

In this paper, we summarize three major pitfalls: (1) \textbf{Data preprocessing discrepancy}. If two EE works conduct evaluations on the same dataset but adopt different preprocessing methods, their results are not directly comparable. Since EE datasets have complex data formats (involving multiple heterogeneous elements including event triggers, arguments, entities, temporal expressions, etc.), data preprocessing methods of existing works often disagree on some design choices, like whether to include multi-token triggers, which results in major data discrepancies. For instance, for the widely-used English subset of ACE 2005~\citep{walker2006ace}, the preprocessing of \citet{wadden-etal-2019-entity} gets $5,055$ event triggers, but \citet{wang-etal-2021-cleve} have $5,349$. 
(2) \textbf{Output space discrepancy}. 
Different model paradigms have inconsistent output spaces, which makes the evaluation metrics of different-paradigm models often not calculated on the same bases. For example, the phrase \textit{Elon Musk} is one argument candidate in the output space of conventional classification-based methods, and it is regarded as one error case when the model misclassifies it. But other model paradigms, like the sequence labeling, have more free output formats and can make two independent predictions for the two tokens \textit{Elon} and \textit{Musk}, which will account for two error cases in the evaluation metric calculation. Larger output spaces of the new model paradigms also result in unclear mappings between predictions and annotations in some cases, which are often overlooked in EE evaluation implementations and lead to problematic results. These details are presented in \cref{sec:output_space}.
(3) \textbf{Absence of pipeline evaluation}.
Recent works handling only the EAE subtask often evaluate the performances based on gold event triggers~\citep{subburathinam-etal-2019-cross,xi-etal-2021-capturing,paie}. In contrast, conventional EE works often conduct pipeline evaluation, i.e., evaluate EAE performances based on triggers predicted at the ED stage. The absence of pipeline evaluation makes these EAE-only works hard to be directly compared with EE works. This has discouraged the research community from considering all the EE subareas in a holistic view. Moreover, only using gold triggers in evaluation cannot evaluate EAE models' resistance to the noise of predicted triggers, which is important in real-world application scenarios.

We conduct systematic meta-analyses of EE papers and empirical experiments, demonstrating the pitfalls' broad and significant influence. We suggest a series of remedies to avoid these pitfalls, including specifying data preprocessing methods, standardizing outputs, and providing pipeline evaluation results. To help conveniently achieve these remedies, we develop a consistent evaluation framework, \textsc{OmniEvent}, which contains implementations for data preprocessing and output standardization, and off-the-shelf predicted triggers on widely-used datasets for easier pipeline evaluation.

To summarize, our contributions are two-fold: (1) We systematically analyze the inconspicuous pitfalls of EE evaluations and demonstrate their significant influence with meta-analyses and experiments. (2) We propose corresponding remedies to avoid the pitfalls and develop a consistent evaluation framework to help implement them.

\section{Related Work}
\label{sec:relate}

Traditional methods~\citep{ji2008refining,gupta-ji:2009:Short,hong-etal-2011-using,li2013joint} rely on human-crafted features and rules to extract events. Most modern EE models automate feature learning with neural networks~\citep{nguyen-grishman-2015-event,nguyen-etal-2016-joint-event,nguyen2018graph} and adopt different model paradigms to model the EE task. The most common \textbf{classification}-based methods view EE as classifying given trigger and argument candidates into different labels~\citep{chen2015event,feng-etal-2016-language,chen-etal-2017-automatically,liu-etal-2018-jointly,wang-etal-2019-adversarial-training,lai-etal-2020-event,wang-etal-2021-cleve,wang-etal-2022-query}. \textbf{Sequence labeling} methods~\citep{nguyen-etal-2016-joint-event,chen-etal-2018-collective,araki-mitamura-2018-open,ding-etal-2019-event,ma-etal-2020-resource,nguyen-etal-2021-crosslingual,guzman-nateras-etal-2022-cross} do EE by labeling every word following a certain tagging schema such as BIO~\citep{ramshaw1999text}. Recently, some works~\citep{du-cardie-2020-event,li-etal-2020-event,liu-etal-2020-event,liu-etal-2021-machine,wei-etal-2021-trigger,sheng-etal-2021-casee,zhou-etal-2022-multi} propose to cast the task formalization of EE into resource-rich machine reading comprehension tasks and adopt the \textbf{span prediction} paradigm to predict the starting and ending positions of event trigger and argument spans. With the development of generative pre-trained language models~\citep{lewis-etal-2020-bart,2020t5,brown2020language}, there have been works~\citep{lu-etal-2021-text2event,xi-etal-2021-capturing,li-etal-2021-document,li-kipt-2022,liu-etal-2022-dynamic,huang-etal-2022-multilingual-generative,du-etal-2022-dynamic,hsu-etal-2022-degree,zeng-etal-2022-ea2e} exploring the \textbf{conditional generation} paradigm to generate sequences indicating EE results.

A few previous works~\citep{wadden-etal-2019-entity,lai-etal-2020-event,wang-etal-2020-maven,wang-etal-2022-query} have noted that data preprocessing discrepancy may influence evaluation results, but they did not especially study its impact with in-depth analyses. To the best of our knowledge, we are the first to study all three kinds of pitfalls of EE evaluation and propose comprehensive remedies for them. 
\section{Pitfalls of Event Extraction Evaluation}
We first introduce our investigation setup for meta-analysis and empirical analysis (\cref{sec:invest_setup}). Then we analyze the three pitfalls: data preprocessing discrepancy (\cref{sec:data_pre_discrepancy}), output space discrepancy (\cref{sec:output_space}), and absence of pipeline evaluation (\cref{sec:absence_pipeline}).

\subsection{Investigation Setup}
\label{sec:invest_setup}

We adopt the following two investigation methods to analyze the influence of the observed pitfalls.

\paragraph{Meta-Analysis}
To comprehensively understand the research status and investigate the potential influence of the evaluation pitfalls, we analyze a broad range of recent EE studies in the meta-analysis. Specifically, we manually retrieve all published papers concerning EE, ED, and EAE tasks at four prestigious venues from 2015 to 2022 via keyword\footnote{We use \textit{event} and \textit{extraction} as keywords for searching. } matching and manual topic rechecking by the authors. The complete paper list is shown in \cref{sec:review}, including $44$ at ACL, $39$ at EMNLP, $19$ at NAACL, and $14$ at COLING. 

We conduct statistical analyses of these papers and their released codes (if any) from multiple perspectives. These statistics will be presented to demonstrate the existence and influence of the pitfalls in the following sections, respectively.

\paragraph{Empirical Analysis}
In addition to the meta-analysis, we conduct empirical experiments to quantitatively analyze the pitfalls' influence on EE evaluation results. We reproduce several representative models covering all four model paradigms mentioned in \cref{sec:relate} to systematically study the influence. Specifically, the models contain: (1) \textbf{Classifcation} methods, including DMCNN~\citep{chen2015event} , DMBERT~\citep{wang-etal-2019-adversarial-training, wang-etal-2019-hmeae}, and CLEVE~\citep{wang-etal-2021-cleve}.
DMCNN and DMBERT adopt
a dynamic multi-pooling operation over hidden representations of convolutional neural networks and BERT~\citep{devlin-etal-2019-bert}, respectively. CLEVE is 
an event-aware pre-trained model enhanced with event-oriented contrastive pre-training.
(2) \textbf{Sequence labeling} methods, including BiLSTM+CRF~\citep{wang-etal-2020-maven} and BERT+CRF~\citep{wang-etal-2020-maven}, which adopt the conditional random field~\citep{crf} as the output layer to make structural predictions.
(3) \textbf{Span prediction} methods, including EEQA~\citep{du-cardie-2020-event} converting EE into a question-answering task, and PAIE~\citep{paie}, which is a prompt-tuning-based EAE method.
(4) \textbf{Conditional generation} method, including Text2Event~\citep{lu-etal-2021-text2event}, which 
is a sequence-to-structure generative EE method with constrained decoding and curriculum learning.

The models are reproduced based on the evaluation settings described in their original papers and released open-source codes (if any). From our meta-analysis, $70$\% of the EE papers adopt the English subset of ACE 2005 dataset~\citep{walker2006ace}\footnote{For brevity, refer to as ``ACE 2005'' in the following.} in their experiments. Hence we also adopt this most widely-used dataset in our empirical experiments to analyze the pitfalls without loss of generality. The reproduction performances are shown in \cref{tab:reproduced}. Following the conventional practice, we report precision (P), recall (R), and the F1 score. In the following analyses, we show the impact of three pitfalls by observing how the performances change after controlling the pitfalls' influence.

\begin{table}[!t]
  \centering
   \small
\begin{adjustbox}{max width=1\linewidth}
{
    \begin{tabular}{l|ccc|ccc}
    \toprule
          & \multicolumn{3}{c|}{\textbf{ED}} & \multicolumn{3}{c}{\textbf{EAE}} \\
    \midrule
    \textbf{Metric} & \textbf{P} & \textbf{R} & \textbf{F1}    & \textbf{P} & \textbf{R} & \textbf{F1} \\
    \midrule
    DMCNN & $65.0$& $69.7$& $67.2$& $45.3$& $41.6$& $43.2$ \\
    DMBERT & $72.1$& $77.1$& $74.5$& $50.5$& $60.0$& $54.8$ \\
    CLEVE & $76.4$& $80.4$& $78.3$& $56.9$& $65.9$& $61.0$ \\
    BiLSTM+CRF & $72.3$& $79.1$& $75.5$& $27.1$& $32.3$& $29.4$  \\
    BERT+CRF & $69.9$& $74.6$& $72.1$& $41.4$& $43.6$& $42.5$ \\
    EEQA & $65.3$& $74.5$& $69.5$& $49.7$& $45.4$& $47.4$ \\
    PAIE & \texttt{N/A} & \texttt{N/A} & \texttt{N/A} & $70.6$& $73.2$& $71.8$ \\
    Text2Event & $66.9$& $72.4$& $69.5$& $48.0$& $54.1$& $50.8$  \\
    \bottomrule
    \end{tabular}
}
\end{adjustbox}
    \caption{The reproduction performances (\%) on ACE 2005 under respective original evaluation settings. ``\texttt{N/A}'' means not applicable as PAIE is an EAE-only model. Reproduction details are introduced in \cref{sec:reproduce_details}}
  \label{tab:reproduced}%
\end{table}

\subsection{Data Preprocessing Discrepancy}
\label{sec:data_pre_discrepancy}

\begin{table*}[!t]
  \centering
   \small
\begin{adjustbox}{max width=1\linewidth}
{
\begin{tabular}{lrrrrrrrr}
\toprule
& \multicolumn{1}{c}{\textbf{Paper\%}} & \multicolumn{1}{c}{\textbf{\#Token}} & \multicolumn{1}{c}{\textbf{\#Trigger}} & \multicolumn{1}{c}{\textbf{\#Argument}} & \multicolumn{1}{c}{\textbf{\#Event Type}} & \multicolumn{1}{c}{\textbf{\#Arg. Role}} & \multicolumn{1}{c}{\textbf{\#Tri. Candidate}} & \multicolumn{1}{c}{\textbf{\#Arg. Candidate}} \\
\midrule
\acedygie   &  $14$  &  $305,266$  & $5,055$  &  $6,040$ & $33$ & $22$  & $305,266$ & $34,474$  \\
\aceoneie   &  $19$  &  $310,020$  & $5,311$  &  $8,055$ & $33$ & $22$ & $309,709$ & $54,650$   \\
\acefull    &  $4$ &  $300,477$  & $5,349$  &  $9,683$   & $33$ & $35$ & $300,165$ & $59,430$   \\
\texttt{Unspecified} &    $63$                         & \multicolumn{1}{r}{-}    & \multicolumn{1}{r}{-}  & \multicolumn{1}{r}{-}    & \multicolumn{1}{r}{-}          & \multicolumn{1}{r}{-}           & \multicolumn{1}{r}{-}                   & \multicolumn{1}{r}{-}      \\ 
\bottomrule            
\end{tabular}
}
\end{adjustbox}
    \caption{The statistics of different ACE 2005 preprocessing scripts. ``Paper\%'' represents the utilization rates
    of different scripts among surveyed papers. \texttt{Unspecified} includes papers ($61\%$) that neither refer to a preprocessing method nor release their preprocessing codes and papers ($2\%$) that release their preprocessing codes that are only used in their own papers. ``Arg.'' and ``Tri.'' are short for argument and trigger, respectively.
    }
  \label{tab:data_pre}
\end{table*}

\begin{table}[t]
    \centering
    \small
\begin{adjustbox}{max width=1\linewidth}
    \begin{tabular}{lccc}
    \toprule
    & \textbf{\acedygie} & \textbf{\aceoneie} & \textbf{\acefull} \\ 
    \midrule
    NLP Toolkit & spaCy & NLTK & CoreNLP \\
    Entity Mention & head & head & full \\
    Multi-token Tri. & \ding{53} & \ding{52} & \ding{52} \\ 
    Temporal Exp. &  \ding{53} &  \ding{53} & \ding{52}\\ 
    Value Exp. &  \ding{53} &  \ding{53} & \ding{52}\\
    Pronoun &  \ding{53} & \ding{52} & \ding{52}\\
    \bottomrule
    \end{tabular}
\end{adjustbox}
    \caption{The major differences between the three preprocessing scripts. NLP Toolkit: the toolkit used for sentence segmentation and tokenization.
    Entity Mention: using head words or full mentions as entity mentions. Multi-token Tri.: whether include multi-token triggers. Temporal Exp., Value Exp., and Pronoun: whether include temporal expressions, values, and pronouns.}
    \label{tab:ace_diff}
\end{table}

Due to the inherent task complexity, EE datasets naturally involve multiple heterogeneous annotation elements. For example, besides event triggers and arguments, EE datasets often annotate entities, temporal expressions, and other spans as argument candidates. The complex data format makes the data preprocessing methods easily differ in many details, which makes the reported results on the same dataset not directly comparable. However, this pitfall has not received extensive attention.

\looseness=-1 To carefully demonstrate the differences brought by data preprocessing discrepancy, we conduct detailed meta-analyses taking the most widely-used ACE 2005 as an example. From all the $116$ surveyed papers, we find three repetitively used open-source preprocessing scripts: \acedygie~\citep{wadden-etal-2019-entity}, \aceoneie~\citep{lin-etal-2020-joint}, and \acefull~\citep{wang-etal-2019-hmeae}.
In addition to these scripts, there are $6$ other open-source preprocessing scripts that are only used once. The utilization rates and data statistics of the different preprocessing methods are shown in \cref{tab:data_pre}. From the statistics, we can observe that: (1) The data differences brought by preprocessing methods are significant. The differences mainly come from the different preprocessing implementation choices, as summarized in \cref{tab:ace_diff}. For instance, \acedygie and \aceoneie ignore the annotated temporal expressions and values in ACE 2005, which results in 13 fewer argument roles compared to \acefull. Intuitively, the significant data discrepancy may result in inconsistent evaluation results.
(2) Each preprocessing script has a certain utilization rate and the majority ($63$\%) papers do not specify their preprocessing methods. The high preprocessing inconsistency and \texttt{Unspecified} rate both show that our community has not fully recognized the significance of the discrepancies resulting from differences in data preprocessing.

\begin{table}[!t]
  \centering
   \small
\begin{adjustbox}{max width=1\linewidth}
{
    \setlength{\tabcolsep}{2pt}
    \begin{tabular}{l|cc|cc|cc}
    \toprule
          & \multicolumn{2}{c|}{\textbf{\acedygie}} & \multicolumn{2}{c|}{\textbf{\aceoneie}} & \multicolumn{2}{c}{\textbf{\acefull}}\\
    \midrule
    \textbf{Metric} & $\Delta$\textbf{ED F1} & $\Delta$\textbf{EAE F1}   & $\Delta$\textbf{ED F1} & $\Delta$\textbf{EAE F1} & $\Delta$\textbf{ED F1} & $\Delta$\textbf{EAE F1}\\
    \midrule
    DMCNN & $-4.7$& $-9.2$& $-4.3$& $-8.0$ & $-$ & $-$ \\
    DMBERT & $-6.3$& $-6.7$& $-5.2$& $-7.6$ & $-$ & $-$ \\
    CLEVE & $-5.4$& $-6.2$& $-3.3$& $-6.3$ & $-$ & $-$ \\
    BiLSTM+CRF & $-3.8$& $+3.1$& $-4.1$& $+3.2$  & $-$ & $-$ \\
    BERT+CRF & $-4.2$& $+2.4$& $-4.2$& $+3.4$ & $-$ & $-$ \\
    EEQA & $-$ & $-$ & $-0.5$& $+0.1$& $+3.6$ & $-4.1$ \\
    PAIE & \texttt{N/A} & $-$ & \texttt{N/A} & $-0.7$ & \texttt{N/A} & $-15.2$ \\
    Text2Event & $-$ & $-$ & $+2.5$& $+3.0$& $+4.7$& $-1.0$  \\
    \bottomrule
    \end{tabular}
}
\end{adjustbox}
    \caption{The F1 (\%) differences between using ACE 2005 preprocessed by another script and the original script. ``$-$'' indicates the model is originally trained and evaluated on this script.}
  \label{tab:exp_datapre}%
\end{table}

To further empirically investigate the influence of preprocessing, we conduct experiments on ACE 2005. \Cref{tab:exp_datapre} shows the F1 differences keeping all settings unchanged except for the preprocessing scripts. We can observe that the influence of different preprocessing methods is significant and varies from different models. 
It indicates that the evaluation results on the same dataset are not necessarily comparable due to the unexpectedly large influence of different preprocessing details.

Moreover, besides ACE 2005, there are also data preprocessing discrepancies in other datasets. 
For example, in addition to the implementation details, the data split of the KBP dataset is not always consistent~\citep{li-treasure-2021, li-kipt-2022}, and some used LDC\footnote{\url{https://www.ldc.upenn.edu/}} datasets are not freely available, such as LDC2015E29. Based on all the above analyses, we suggest the community pay more attention to data discrepancies caused by preprocessing, and we propose corresponding remedies in \cref{sec:specify_preprocess}.

\begin{figure*}[!ht]
  \centering
  \includegraphics[width=\linewidth]{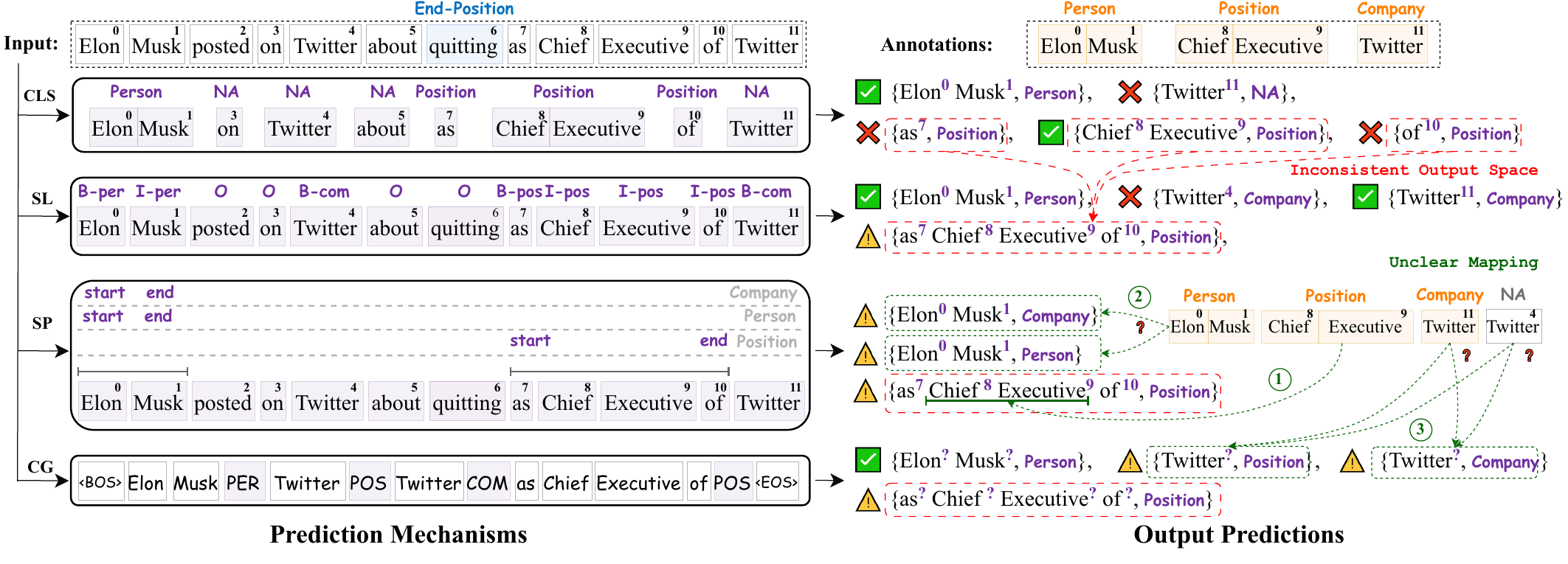}
  \caption{An illustration for the prediction mechanisms of different model paradigms and their corresponding output formats in event argument extraction task. Classification (CLS) methods do multi-class classification on the argument candidates within a pre-defined set. Sequence labeling (SL) methods predict a  BIO-label for each input token. Span prediction (SP) models predict the starting and ending indices of a span for each argument role. Conditional generation (CG) models directly generate a structured sequence consisting of argument mentions and roles. Predictions marked with yellow warning signs are tricky samples for pitfalls illustration. Best viewed in color. }
  \label{fig:output_space}
\end{figure*}

\begin{figure}[!t]
  \centering
  \includegraphics[width=0.80\linewidth]{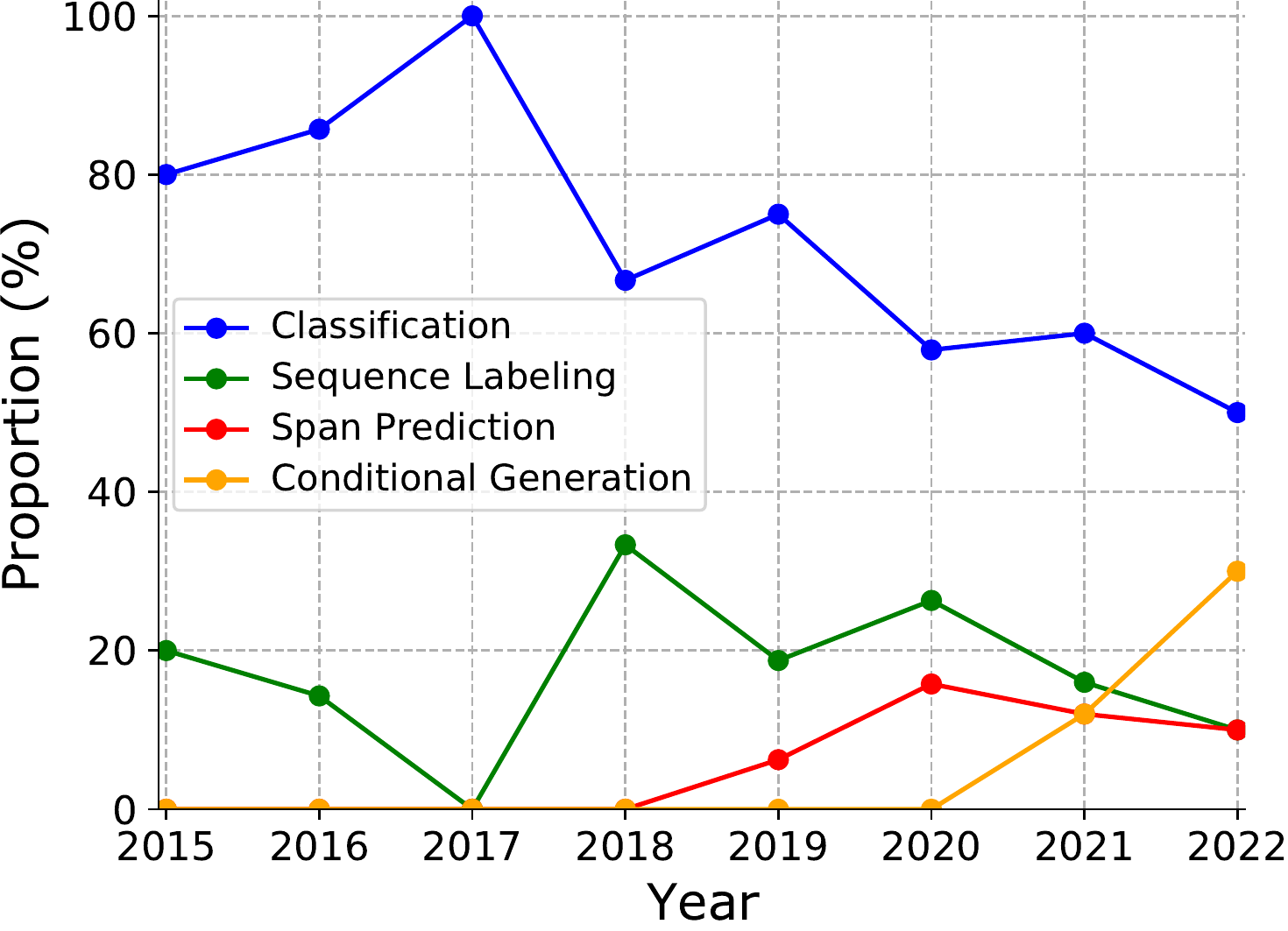}
  \caption{Proportion per year of EE papers adopting different model paradigms from 2015 to 2022.}
  \label{fig:paradigm_num}
\end{figure}

\subsection{Output Space Discrepancy}
\label{sec:output_space}

As shown in \cref{fig:paradigm_num}, the diversity of adopted model paradigms in EE studies has substantially increased in recent years. \cref{fig:output_space} illustrates the different paradigms' workflows in EAE scenario\footnote{The ED workflows are the same as EAE or even simpler.}. The paradigms inherently have very different output spaces, which results in inconspicuous pitfalls in the comparative evaluations across paradigms.

\paragraph{Inconsistent Output Spaces between Different Paradigms}

As shown in \cref{fig:output_space}, there are substantial differences between the model output spaces of different paradigms. CLS-paradigm models only output a unique label for each candidate in a pre-defined set. While models of SL and SP paradigms can make predictions for any consecutive spans in the input sequence. The output space of CG-paradigm models is even larger, as their vanilla\footnote{Indicates excluding tricks like vocabulary constraint, etc.} output sequences are completely free, e.g., they can even involve tokens unseen in the input. The inconsistent output spaces make the evaluation metrics of different-paradigm models calculated on different bases and not directly comparable. For instance, when calculating the confusion matrices for the prediction \textit{as Chief Executive of} in \cref{fig:output_space}, the CLS paradigm takes it as one true positive (TP) and two false positives (FP), while the remaining paradigms only count it as one FP. The CLS paradigm may also have an advantage in some cases since it is constrained by the pre-defined candidate sets and cannot make illegal predictions as other paradigms may have.

\paragraph{Unclear Mappings between Predictions and Annotations}

Implementing the mappings between model predictions and dataset annotations is a key component for evaluation. The larger output spaces of SL, SP, and CG paradigms often produce unclear mappings, which are easily neglected in the EE evaluation implementations and influence the final metrics. As shown in \cref{fig:output_space} (bottom right), we summarize three major unclear mapping issues:
\textcolor[RGB]{6,104,0}{\normalsize{\textcircled{\scriptsize{\textbf{1}}}}} \textbf{Prediction span overlaps the gold span.} A prediction span of non-CLS paradigm models may overlap but not strictly align with the annotated span, bringing in an unclear implementation choice. As in \cref{fig:output_space}, it is unclear whether the predicted role \texttt{Position} for the span \textit{as Chief Executive of} should be regarded as a correct prediction for the contained annotated span \textit{Chief Executive}.
\textcolor[RGB]{6,104,0}{\normalsize{\textcircled{\scriptsize{\textbf{2}}}}} \textbf{Multiple predictions for one annotated span.} If without special constraints, models of SP and CG paradigms may make multiple predictions for one span.
\cref{fig:output_space} presents two contradictory predictions (\texttt{Company} and \texttt{Person}) for the annotated span \textit{Elon Musk}. To credit the correct one only or penalize both should lead to different evaluation results.
\textcolor[RGB]{6,104,0}{\normalsize{\textcircled{\scriptsize{\textbf{3}}}}} \textbf{Predictions without positions for non-unique spans.} Vanilla CG-paradigm models make predictions by generating contents without specifying their positions. When the predicted spans are non-unique in the inputs, it is unclear how to map them to annotated spans in different positions. As in \cref{fig:output_space}, the CG model outputs two \textit{Twitter} predictions, which can be mapped to two different input spans.

\looseness=-1 To quantitatively demonstrate the influence of output space discrepancy, we conduct empirical experiments. Specifically, we propose an output standardization method (details in {\cref{sec:standardize_output}}), which unify the output spaces of different paradigms and handle all the unclear mapping issues. We report the changes in metrics between the original evaluation implementations and the evaluation with our output standardization in \cref{tab:exp_output}. We can see the results change obviously, with the maximum increase and decrease of $+2.8$ in ED precision and $-3.5$ in EAE recall, respectively. It indicates the output space discrepancy can lead to highly inconsistent evaluation results. Hence, we advocate for awareness of the output space discrepancy in evaluation implementations and suggest doing output standardization when comparing models using different paradigms.

\begin{table}[!t]
  \centering
   \small
\begin{adjustbox}{max width=1\linewidth}
{
    \begin{tabular}{l|ccc|ccc}
    \toprule
          & \multicolumn{3}{c|}{\textbf{ED}} & \multicolumn{3}{c}{\textbf{EAE}} \\
    \midrule
    \textbf{Metric} & $\Delta$\textbf{P} & $\Delta$\textbf{R} & $\Delta$\textbf{F1} 
    & $\Delta$\textbf{P} & $\Delta$\textbf{R} & $\Delta$\textbf{F1} \\    
    \midrule
    BiLSTM+CRF & $+2.0$& $-0.1$& $+1.0$& $+5.0$& $-0.3$& $+2.3$\\
    BERT+CRF & $+2.8$& $-0.2$& $+1.3$& $+2.6$& $-0.3$& $+1.2$  \\
    EEQA & $-0.6$& $+0.8$& $+0.1$& $-3.1$& $-2.1$& $-2.6$ \\
    PAIE & \texttt{N/A} & \texttt{N/A} & \texttt{N/A} & $+4.6$& $-0.6$& $+2.0$  \\
    Text2Event & $+1.1$& $+0.0$& $+0.6$& $-1.1$& $-3.5$& $-2.3$ \\
    \bottomrule
    \end{tabular}
}
\end{adjustbox}
    \caption{The precision, recall, and F1 (\%) differences between evaluation with and without our output standardization. The results are evaluated on \aceoneie, and the results for other preprocessing methods are in \cref{sec:app_add_exp}. Output standardization aligns the output spaces of the other paradigms into that of the CLS paradigm, and thus we do not include the CLS-paradigm models here, whose results are unchanged. }
  \label{tab:exp_output}
\end{table}

\subsection{Absence of Pipeline Evaluation}

\label{sec:absence_pipeline}
The event extraction (EE) task is typically formalized as a two-stage pipeline, i.e., first event detection (ED) and then event argument extraction (EAE). In real applications, EAE is based on ED and only extracts arguments for triggers detected by 
the ED model. Therefore, the conventional evaluation of EAE is based on predicted triggers and considers ED prediction errors, which we call \textbf{pipeline evaluation}.
It assesses the overall performance of an event extraction system and is consistent with real-world pipeline application scenarios. 

\begin{figure}
  \centering
  \includegraphics[width=0.80\linewidth]{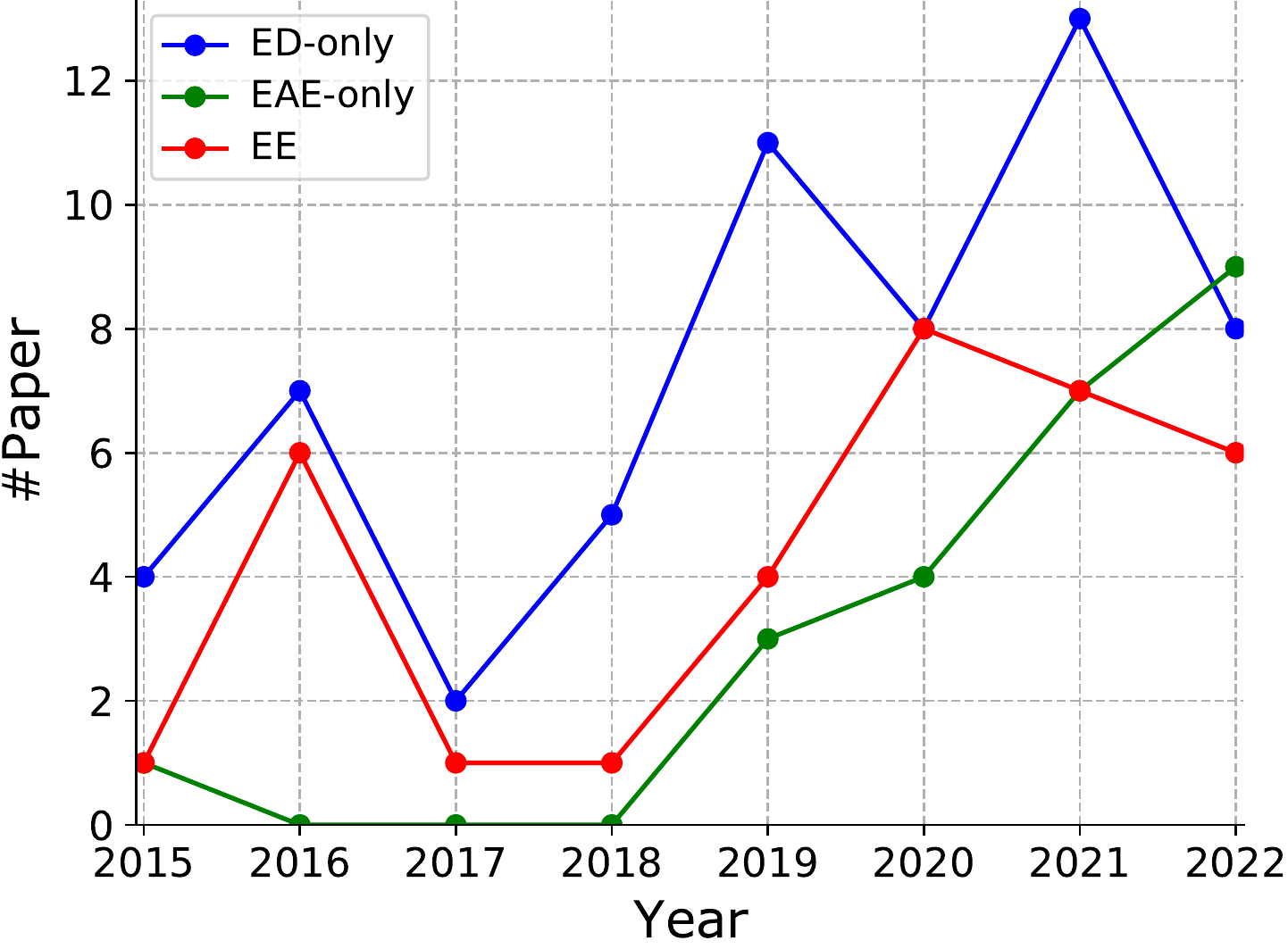}
  \caption{Number of papers concerning only ED, only EAE, and EE tasks from 2015 to 2022.}
  \label{fig:eae_proportion}
\end{figure}

\looseness=-1 However, as shown in \cref{fig:eae_proportion}, more and more works have focused only on EAE in recent years. For convenience and setting a unified evaluation base between the EAE-only works, $95.45$\% of them only evaluate EAE taking gold triggers as inputs. We dub this evaluation setting as \textbf{gold trigger evaluation}. The conventional pipeline evaluation of EE works is absent in most EAE-only works, which poses two issues: (1) The absence of pipeline evaluation makes the results of EAE-only works hard to be directly cited and compared in EE studies. In the papers covered by our meta-analysis, there is nearly no direct comparison between EE methods and EAE-only methods. It indicates that the evaluation setting difference has created a gap between the two closely connected research tasks, which hinders the community from comprehensively understanding the research status. (2) The gold trigger evaluation may not well reflect the real-world performance since it ignores the EAE models' resistance to trigger noise.
In real-world applications, the input triggers for EAE models are noisy predicted triggers. A good EAE method should be resistant to trigger noise, e.g., not extracting arguments for false positive triggers. The gold trigger evaluation neglects trigger noise. 
\begin{table}[!t]
  \centering
   \small
\begin{adjustbox}{max width=1\linewidth}
{
    \begin{tabular}{l|c|c|c}
    \toprule
    \multirow{2}{*}{\textbf{Metric}}  & \multirow{2}{*}{\textbf{ED F1}}  & \multicolumn{1}{c|}{\textbf{Gold Tri.}} & \multicolumn{1}{c}{\textbf{Pipeline}} \\
     &  & \textbf{EAE F1} &\textbf{EAE F1} \\    \midrule
    DMCNN & $62.8$ &  $51.6$ & $35.2$  \\
    DMBERT & $69.4$ &  $67.2$ & $47.2$  \\
    CLEVE & $75.0$ & $69.6$ & $54.7$ \\
    BiLSTM+CRF & $72.4$  & $45.3$ & $34.9$ \\
    BERT+CRF & $69.2$  & $64.3$ & $47.1$  \\
    EEQA & $69.1$ & $63.9$  & $45.0$ \\
    PAIE & $75.0$ &  $73.2$ & $56.7$  \\
    \bottomrule
    \end{tabular}
}
\end{adjustbox}
    \caption{EAE F1 scores (\%) of gold trigger evaluation 
    and pipeline evaluation on \aceoneie. Results for other preprocessing methods are in \cref{sec:app_add_exp}. We also report corresponding ED F1 scores to show trigger quality. PAIE adopts the triggers predicted by CLEVE. The joint model Text2Event is excluded since its trigger input cannot be controlled.
    }
  \label{tab:exp_pipeline}
\end{table}

To assess the potential influence of this pitfall, we compare experimental results under the gold trigger evaluation and pipeline evaluation of various models in \cref{tab:exp_pipeline}. We can observe different trends from the results of gold trigger evaluation and pipeline evaluation. For example, although DMBERT performs much better than BERT+CRF under gold trigger evaluation, they perform nearly the same under pipeline evaluation ($47.2$ vs. $47.1$). It suggests that the absence of pipeline evaluation may bring obvious result divergence, which is rarely noticed in existing works. Based on the above discussions, we suggest also conducting the pipeline evaluation in EAE works.

\section{Consistent Evaluation Framework}
\begin{table*}[!t]
  \centering
  \small
  \begin{adjustbox}{max width=1\linewidth}
  {
    \begin{tabular}{l|ccc|ccc|ccc|ccc}
    \toprule
    & \multicolumn{6}{c|}{\textbf{Original Evaluation}} & \multicolumn{6}{c}{\textbf{Consistent Evaluation}} \\
    \cmidrule{2-13}
    & \multicolumn{3}{c|}{\textbf{ED}} & \multicolumn{3}{c|}{\textbf{EAE}}
     & \multicolumn{3}{c|}{\textbf{ED}} & \multicolumn{3}{c}{\textbf{EAE}} \\ 
     \cmidrule{1-13}
     \textbf{Metric} & \textbf{P} & \textbf{R} & \textbf{F1} & \textbf{P} & \textbf{R} & \textbf{F1} & \textbf{P} & \textbf{R} & \textbf{F1} & \textbf{P} & \textbf{R} & \textbf{F1} \\
     \midrule
     DMCNN & $75.6$& $63.6$& $69.1$& $62.2$& $46.9$& $53.5$ & $65.0$& $69.7$& $67.2$& $45.3$& $41.6$& $43.2$ \\
     DMBERT & $77.6$& $71.8$& $74.6$& $58.8$& $55.8$& $57.2$ & $72.1$& $77.1$& $74.5$& $50.5$& $60.0$& $54.8$ \\
     CLEVE & $78.1$& $81.5$& $79.8$& $55.4$& $68.0$& $61.1$ & $76.4$& $80.4$& $78.3$& $56.9$& $65.9$& $61.0$ \\
     BiLSTM+CRF & $77.2$ & $74.9$ & $75.4$ & $27.1^*$ & $32.3^*$ & $29.5^*$ & $74.2$& $78.9$& $76.5$& $42.8$& $32.4$& $36.9$ \\
     BERT+CRF & $71.3$& $77.1$& $74.1$& $41.4^*$& $43.6^*$& $42.5^*$ & $72.4$& $74.5$& $73.4$& $55.6$& $43.2$& $48.6$ \\
     EEQA & $71.1$& $73.7$& $72.4$& $56.9$& $49.8$& $53.1$ & $70.5$& $77.3$& $73.6$& $65.8$& $25.5$& $36.4$ \\
     PAIE & \texttt{N/A} & \texttt{N/A} & \texttt{N/A} & $70.6^*$& $73.2^*$& $72.7$ & \texttt{N/A} & \texttt{N/A} & \texttt{N/A} & $61.4$& $46.2$& $52.7$ \\
     Text2Event & $69.6$& $74.4$& $71.9$& $52.5$& $55.2$& $53.8$ & $76.1$& $74.5$& $75.2$& $59.6$& $43.0$& $50.0$ \\
    \bottomrule
    \end{tabular}
  }
  \end{adjustbox}
  \caption{Experimental results (\%) under the original evaluation and our consistent evaluation on \acefull 
  preprocessed dataset. The ``original evaluation'' results are directly taken from respective original papers, except the $^*$ results, which are missed in the original papers and from our reproduction. All the results are under pipeline evaluation, except for the ``original evaluation'' results of PAIE, which originally adopts the gold trigger evaluation. Experimental results for other preprocessing methods are in \cref{sec:app_add_exp}.
  }
  \label{tab:main_exp}
\end{table*}

The above analyses show that the hidden pitfalls substantially harm the consistency and validity of EE evaluation. We propose a series of remedies to avoid these pitfalls and develop a consistent evaluation framework, \textsc{OmniEvent}. \textsc{OmniEvent} helps to achieve the remedies and eases users of handling the inconspicuous preprocessing and evaluation details. It is publicly released and continually maintained to handle emerging evaluation pitfalls. The suggested remedies include specifying data preprocessing (\cref{sec:specify_preprocess}), standardizing outputs (\cref{sec:standardize_output}), and providing pipeline evaluation results (\cref{sec:pipeline_evaluation}). We further re-evaluate various EE models using our framework and analyze the results in \cref{sec:experimental_results}.

\subsection{Specify Data Preprocessing}
\label{sec:specify_preprocess}
As analyzed in \cref{sec:data_pre_discrepancy}, preprocessing discrepancies have an obvious influence on evaluation results. The research community should pay more attention to data preprocessing details and try to specify them. Specifically, we suggest future EE works adopt a consistent preprocessing method on the same dataset. Regarding the example in \cref{sec:data_pre_discrepancy}, for the multiple ACE 2005 preprocessing scripts, we recommend \acefull since it retains the most comprehensive event annotations, e.g., multi-token triggers and the time-related argument roles, which are commonly useful in real-world applications. If a study has to use different preprocessing methods for special reasons, we suggest specifying the preprocessing method with reference to public codes. However, there are no widely-used publicly available preprocessing scripts for many EE datasets, which makes many researchers have to re-develop their own preprocessing methods.
In our consistent evaluation framework, we provide preprocessing scripts for various widely-used datasets, including ACE 2005~\citep{walker2006ace}, TAC KBP Event Nugget Data 2014-2016~\citep{ellis2014overview, ellis2015overview, ellis2016overview}, TAC KBP 2017~\citep{getman2017overview}, RichERE~\citep{song2015light}, MAVEN~\citep{wang-etal-2020-maven}, LEVEN~\citep{leven}, DuEE~\citep{duee}, and FewFC~\citep{zhou2021role}. We will continually add the support of more datasets, such as RAMS~\citep{ebner2019multi} and WikiEvents~\citep{li-etal-2021-document}, and we welcome the community to contribute scripts for more datasets.

\subsection{Standardize Outputs}
\label{sec:standardize_output}
Based on the discussions about output space discrepancy in \cref{sec:output_space}, we propose and implement an output standardization method in our framework. 

To mitigate the inconsistency of output spaces between paradigms, we project the outputs of non-CLS paradigm models onto the most strict CLS-paradigm output space. Specifically, we follow strict boundary-matching rules to assign the non-CLS predictions to each trigger/argument candidate in pre-defined candidate sets of the CLS paradigm. The final evaluation metrics are computed purely on the candidate sets, and those predictions that fail to be matched are discarded. The intuition behind this operation is that given the CLS-paradigm candidate sets are automatically constructed, the illegal predictions out of this scope can also be automatically filtered in real-world applications.

Regarding the unclear mappings between predictions and annotations, we consider the scenario of real-world applications and propose several deterministic mapping rules for consistent evaluations. We respond to the issues mentioned in \cref{sec:output_space} as follows.
\textcolor[RGB]{6,104,0}{\normalsize{\textcircled{\scriptsize{\textbf{1}}}}} \textbf{Prediction span overlaps the gold span.} We follow strict boundary-matching rules and discard such overlapping predictions. For example, the SL prediction of \textit{as Chief Executive of} cannot strictly match any candidate in the candidate set of the CLS paradigm. Hence it is discarded after output standardization. \textcolor[RGB]{6,104,0}{\normalsize{\textcircled{\scriptsize{\textbf{2}}}}} \textbf{Multiple predictions for one annotated span.} If the outputs are with confidence scores, we choose the prediction with the highest confidence as the final prediction, otherwise, we simply choose the first appearing prediction. The remaining predictions are discarded.  \textcolor[RGB]{6,104,0}{\normalsize{\textcircled{\scriptsize{\textbf{3}}}}} \textbf{Predictions without positions for non-unique spans.} We assign such predictions to the annotated spans simply by their appearing order in the output/input sequence to avoid information leakage. 
We encourage designing new models or post-processing rules to add positional information for CG predictions so that this issue can be directly solved by strict boundary-matching.

\subsection{Provide Pipeline Evaluation Results}
\label{sec:pipeline_evaluation}

The absence of pipeline evaluation (\cref{sec:absence_pipeline}) creates a gap between EE and EAE works, and may not well reflect EAE models' performance in real-world scenarios. Therefore, in addition to the common gold trigger evaluation results, we suggest future EAE-only works also provide pipeline evaluation results. However, there are two difficulties: (1) It is an extra overhead for the EAE-only works to implement an ED model and get predicted triggers on the datasets. (2) If two EAE models use different predicted triggers, their evaluation results are not directly comparable since the trigger quality influences EAE performance. To alleviate these difficulties, our consistent evaluation framework releases off-the-shelf predicted triggers for the widely-used EE datasets, which will help future EAE works conduct easy and consistent pipeline evaluations. The released predicted triggers are generated with existing top-performing ED models so that the obtained pipeline evaluation results shall help the community to understand the possible EE performance of combining top ED and EAE models.

\subsection{Experimental Results}
\label{sec:experimental_results}

\looseness=-1 We re-evaluate various EE models with our consistent evaluation framework. The results are shown in \cref{tab:main_exp}, and we can observe that: (1) If we are not aware of the pitfalls of EE evaluation, we can only understand EE development status and compare competing models from the ``Original Evaluation'' results in \cref{tab:main_exp}. After eliminating the influence of the pitfalls with our framework, the consistent evaluation results change a lot in both absolute performance levels and relative model rankings. This comprehensively demonstrates the influence of the three identified evaluation pitfalls on EE research and highlights the importance of awareness of these pitfalls. Our framework can help avoid the pitfalls and save efforts in handling intensive evaluation implementation details. (2) Although the changes in F1 scores are minor for some models (e.g., CLEVE), their precision and recall scores vary significantly. In these cases, consistent evaluation is also necessary since real-world applications may have different precision and recall preferences.

\section{Conclusion and Future Work}
In this paper, we identify three pitfalls of event extraction evaluation, which are data preprocessing discrepancy, output space discrepancy, and absence of pipeline evaluation. Meta-analyses and empirical experiments present a huge impact of these pitfalls, which urges the attention of our research community. To avoid the pitfalls, we suggest a series of remedies, including specifying data preprocessing, standardizing outputs, and providing pipeline evaluation results. We develop a consistent evaluation framework \textsc{OmniEvent}, to help future works implement these remedies. In the future, we will continually maintain it to well handle more emerging EE datasets, model paradigms, and other possible hidden evaluation pitfalls.

\section*{Limitations}
The major limitations of our work are three-fold: (1) In the empirical experiments, we only train and evaluate models on English datasets. As the analyzed pitfalls are essentially language-independent, we believe the empirical conclusions could generalize to other languages. The developed consistent evaluation framework now includes multiple English and Chinese datasets, and we will extend it to support more languages in the future. 
(2) The three pitfalls analyzed in this paper are identified from our practical experiences and may not cover all the pitfalls of EE evaluation. We encourage the community to pay more attention to finding other possible hidden pitfalls of EE evaluation. We will also continually maintain the proposed consistent evaluation framework to support mitigating the influence of newly-found pitfalls. 
(3) Our meta-analysis only covers papers published at ACL, EMNLP, NAACL, and COLING on mainstream EE research since 2015. Although we believe that we can obtain representative observations from the $116$ surveyed papers, some EE works published at other venues and at earlier times are missed.

\section*{Ethical Considerations}
We discuss the ethical considerations and broader impact of this work here:
(1) \textbf{Intellectual property}. The copyright of ACE 2005 belongs to LDC\footnote{\url{https://www.ldc.upenn.edu/}\label{LDC}}. We access it through our LDC membership and strictly adhere to its license. We believe the established ACE 2005 dataset is desensitized. In our consistent evaluation framework, we will only provide preprocessing scripts rather than preprocessed datasets for those datasets whose licenses do not permit redistribution. The \acedygie preprocessing script\footnote{\url{https://github.com/dwadden/dygiepp}} and the used code repositories for DMCNN\footnote{\url{https://github.com/THU-KEG/MAVEN-dataset}\label{maven-1}}, DMBERT\textsuperscript{\ref{maven-1}}, BiLSTM+CRF\textsuperscript{\ref{maven-1}}, BERT+CRF\textsuperscript{\ref{maven-1}}, EEQA\footnote{\url{https://github.com/xinyadu/eeqa}}, and Text2Event\footnote{\url{https://github.com/luyaojie/Text2Event}} are released under MIT license\footnote{\url{https://opensource.org/licenses/MIT}}. 
These are all public research resources. We use them for the research purpose in this work, which is consistent with their intended use.
(2) \textbf{Intended use}. Our consistent evaluation framework implements the suggested remedies to avoid the identified pitfalls in EE evaluation. Researchers are supposed to use this framework to conduct consistent evaluations for comparing various competing EE models.
(3) \textbf{Misuse risks}. 
The results reported in this paper and the evaluation results produced by our consistent evaluation framework \textbf{should not} be used for offensive arguments or interpreted as implying misconduct of other works. The analyzed pitfalls in this work are inconspicuous and very easy to be accidentally overlooked. Hence the community is generally unaware of them or underestimates their influence. The contribution of our work lies in raising awareness of the pitfalls and helping to avoid them in future works. 
(4) \textbf{Accessibility}. Many widely-used datasets (such as ACE 2005, KBP, etc.) are not freely available to everyone. The financial fairness issue may influence the broader usage of the data for EE research. 
\bibliography{custom}

\begin{thebibliography}{143}
\expandafter\ifx\csname natexlab\endcsname\relax\def\natexlab#1{#1}\fi

\bibitem[{Ahn(2006)}]{ahn2006stages}
David Ahn. 2006.
\newblock \href {http://aclweb.org/anthology/W06-0901} {The stages of event
  extraction}.
\newblock In \emph{Proceedings of ACL Workshop on Annotating and Reasoning
  about Time and Events}, pages 1--8.

\bibitem[{Araki and Mitamura(2018)}]{araki-mitamura-2018-open}
Jun Araki and Teruko Mitamura. 2018.
\newblock \href {https://www.aclweb.org/anthology/C18-1075} {Open-domain event
  detection using distant supervision}.
\newblock In \emph{Proceedings of COLING}, pages 878--891.

\bibitem[{Bekoulis et~al.(2019)Bekoulis, Deleu, Demeester, and
  Develder}]{bekoulis-etal-2019-sub}
Giannis Bekoulis, Johannes Deleu, Thomas Demeester, and Chris Develder. 2019.
\newblock \href {https://doi.org/10.18653/v1/N19-1081} {Sub-event detection
  from twitter streams as a sequence labeling problem}.
\newblock In \emph{Proceedings of NAACL-HLT}, pages 745--750.

\bibitem[{Bronstein et~al.(2015)Bronstein, Dagan, Li, Ji, and
  Frank}]{bronstein-etal-2015-seed}
Ofer Bronstein, Ido Dagan, Qi~Li, Heng Ji, and Anette Frank. 2015.
\newblock \href {https://doi.org/10.3115/v1/P15-2061} {Seed-based event trigger
  labeling: {H}ow far can event descriptions get us?}
\newblock In \emph{Proceedings of ACL-IJCNLP}, pages 372--376.

\bibitem[{Brown et~al.(2020)Brown, Mann, Ryder, Subbiah, Kaplan, Dhariwal,
  Neelakantan, Shyam, Sastry, Askell et~al.}]{brown2020language}
Tom Brown, Benjamin Mann, Nick Ryder, Melanie Subbiah, Jared~D Kaplan, Prafulla
  Dhariwal, Arvind Neelakantan, Pranav Shyam, Girish Sastry, Amanda Askell,
  et~al. 2020.
\newblock \href
  {https://papers.nips.cc/paper/2020/file/1457c0d6bfcb4967418bfb8ac142f64a-Paper.pdf}
  {Language models are few-shot learners}.
\newblock In \emph{Proceedings of NeurIPS}, volume~33, pages 1877--1901.

\bibitem[{Cao et~al.(2022)Cao, Li, Su, Li, Fei, Wu, Li, Zhao, and
  Ji}]{cao-etal-2022-oneee}
Hu~Cao, Jingye Li, Fangfang Su, Fei Li, Hao Fei, Shengqiong Wu, Bobo Li, Liang
  Zhao, and Donghong Ji. 2022.
\newblock \href {https://aclanthology.org/2022.coling-1.170} {{O}ne{EE}: {A}
  one-stage framework for fast overlapping and nested event extraction}.
\newblock In \emph{Proceedings of {COLING}}, pages 1953--1964.

\bibitem[{Cao et~al.(2020)Cao, Chen, Zhao, and
  Wang}]{cao-etal-2020-incremental}
Pengfei Cao, Yubo Chen, Jun Zhao, and Taifeng Wang. 2020.
\newblock \href {https://doi.org/10.18653/v1/2020.emnlp-main.52} {Incremental
  event detection via knowledge consolidation networks}.
\newblock In \emph{Proceedings of EMNLP}, pages 707--717.

\bibitem[{Chan et~al.(2019)Chan, Fasching, Qiu, and Min}]{chan-etal-2019-rapid}
Yee~Seng Chan, Joshua Fasching, Haoling Qiu, and Bonan Min. 2019.
\newblock \href {https://doi.org/10.18653/v1/P19-3006} {Rapid customization for
  event extraction}.
\newblock In \emph{Proceedings of ACL: System Demonstrations}, pages 31--36.

\bibitem[{Chen et~al.(2021)Chen, Lin, Han, and Sun}]{chen-etal-2021-honey}
Jiawei Chen, Hongyu Lin, Xianpei Han, and Le~Sun. 2021.
\newblock \href {https://doi.org/10.18653/v1/2021.emnlp-main.637} {Honey or
  poison? {S}olving the trigger curse in few-shot event detection via causal
  intervention}.
\newblock In \emph{Proceedings of EMNLP}, pages 8078--8088.

\bibitem[{Chen et~al.(2017)Chen, Liu, Zhang, Liu, and
  Zhao}]{chen-etal-2017-automatically}
Yubo Chen, Shulin Liu, Xiang Zhang, Kang Liu, and Jun Zhao. 2017.
\newblock \href {https://doi.org/10.18653/v1/P17-1038} {{Automatically Labeled
  Data Generation for Large Scale Event Extraction}}.
\newblock In \emph{Proceedings of ACL}, pages 409--419.

\bibitem[{Chen et~al.(2015)Chen, Xu, Liu, Zeng, and Zhao}]{chen2015event}
Yubo Chen, Liheng Xu, Kang Liu, Daojian Zeng, and Jun Zhao. 2015.
\newblock \href {https://doi.org/10.3115/v1/P15-1017} {Event extraction via
  dynamic multi-pooling convolutional neural networks}.
\newblock In \emph{Proceedings of ACL-IJCNLP}, pages 167--176.

\bibitem[{Chen et~al.(2018)Chen, Yang, Liu, Zhao, and
  Jia}]{chen-etal-2018-collective}
Yubo Chen, Hang Yang, Kang Liu, Jun Zhao, and Yantao Jia. 2018.
\newblock \href {https://doi.org/10.18653/v1/D18-1158} {Collective event
  detection via a hierarchical and bias tagging networks with gated multi-level
  attention mechanisms}.
\newblock In \emph{Proceedings of EMNLP}, pages 1267--1276.

\bibitem[{Cong et~al.(2021)Cong, Cui, Yu, Liu, Yubin, and
  Wang}]{cong-etal-2021-shot}
Xin Cong, Shiyao Cui, Bowen Yu, Tingwen Liu, Wang Yubin, and Bin Wang. 2021.
\newblock \href {https://doi.org/10.18653/v1/2021.findings-acl.3} {{F}ew-{S}hot
  {E}vent {D}etection with {P}rototypical {A}mortized {C}onditional {R}andom
  {F}ield}.
\newblock In \emph{Findings of ACL-IJCNLP}, pages 28--40.

\bibitem[{Cui et~al.(2020)Cui, Yu, Liu, Zhang, Wang, and
  Shi}]{cui-etal-2020-edge}
Shiyao Cui, Bowen Yu, Tingwen Liu, Zhenyu Zhang, Xuebin Wang, and Jinqiao Shi.
  2020.
\newblock \href {https://doi.org/10.18653/v1/2020.findings-emnlp.211}
  {Edge-enhanced graph convolution networks for event detection with syntactic
  relation}.
\newblock In \emph{Findings of EMNLP}, pages 2329--2339.

\bibitem[{Deng et~al.(2021)Deng, Zhang, Li, Hui, Huaixiao, Chen, Huang, and
  Chen}]{deng-etal-2021-ontoed}
Shumin Deng, Ningyu Zhang, Luoqiu Li, Chen Hui, Tou Huaixiao, Mosha Chen, Fei
  Huang, and Huajun Chen. 2021.
\newblock \href {https://doi.org/10.18653/v1/2021.acl-long.220} {{O}nto{ED}:
  {L}ow-resource event detection with ontology embedding}.
\newblock In \emph{Proceedings of ACL-IJCNLP}, pages 2828--2839.

\bibitem[{Devlin et~al.(2019)Devlin, Chang, Lee, and
  Toutanova}]{devlin-etal-2019-bert}
Jacob Devlin, Ming-Wei Chang, Kenton Lee, and Kristina Toutanova. 2019.
\newblock \href {https://doi.org/10.18653/v1/N19-1423} {{BERT}: {Pre-training
  of Deep Bidirectional Transformers for Language Understanding}}.
\newblock In \emph{Proceedings of NAACL-HLT}, pages 4171--4186.

\bibitem[{Ding et~al.(2019)Ding, Li, Liu, Zheng, and
  Lin}]{ding-etal-2019-event}
Ning Ding, Ziran Li, Zhiyuan Liu, Haitao Zheng, and Zibo Lin. 2019.
\newblock \href {https://doi.org/10.18653/v1/D19-1033} {Event detection with
  trigger-aware lattice neural network}.
\newblock In \emph{Proceedings of EMNLP-IJCNLP}, pages 347--356.

\bibitem[{Du and Cardie(2020{\natexlab{a}})}]{du-cardie-2020-document}
Xinya Du and Claire Cardie. 2020{\natexlab{a}}.
\newblock \href {https://doi.org/10.18653/v1/2020.acl-main.714} {Document-level
  event role filler extraction using multi-granularity contextualized
  encoding}.
\newblock In \emph{Proceedings of ACL}, pages 8010--8020.

\bibitem[{Du and Cardie(2020{\natexlab{b}})}]{du-cardie-2020-event}
Xinya Du and Claire Cardie. 2020{\natexlab{b}}.
\newblock \href {https://doi.org/10.18653/v1/2020.emnlp-main.49} {Event
  extraction by answering (almost) natural questions}.
\newblock In \emph{Proceedings of EMNLP}, pages 671--683.

\bibitem[{Du et~al.(2022)Du, Li, and Ji}]{du-etal-2022-dynamic}
Xinya Du, Sha Li, and Heng Ji. 2022.
\newblock \href {https://doi.org/10.18653/v1/2022.acl-long.361} {Dynamic global
  memory for document-level argument extraction}.
\newblock In \emph{Proceedings of ACL}, pages 5264--5275.

\bibitem[{Ebner et~al.(2020)Ebner, Xia, Culkin, Rawlins, and
  Van~Durme}]{ebner2019multi}
Seth Ebner, Patrick Xia, Ryan Culkin, Kyle Rawlins, and Benjamin Van~Durme.
  2020.
\newblock \href {https://doi.org/10.18653/v1/2020.acl-main.718} {Multi-sentence
  argument linking}.
\newblock In \emph{Proceedings of ACL}, pages 8057--8077.

\bibitem[{Ellis et~al.(2015)Ellis, Getman, Fore, Kuster, Song, Bies, and
  Strassel}]{ellis2015overview}
Joe Ellis, Jeremy Getman, Dana Fore, Neil Kuster, Zhiyi Song, Ann Bies, and
  Stephanie~M Strassel. 2015.
\newblock \href
  {https://www.ldc.upenn.edu/sites/www.ldc.upenn.edu/files/tackbp2015_overview.pdf}
  {Overview of linguistic resources for the {TAC} {KBP} 2015 evaluations:
  {M}ethodologies and results.}
\newblock In \emph{TAC}.

\bibitem[{Ellis et~al.(2016)Ellis, Getman, Fore, Kuster, Song, Bies, and
  Strassel}]{ellis2016overview}
Joe Ellis, Jeremy Getman, Dana Fore, Neil Kuster, Zhiyi Song, Ann Bies, and
  Stephanie~M Strassel. 2016.
\newblock \href
  {https://www.ldc.upenn.edu/sites/www.ldc.upenn.edu/files/tackbp2016-linguistic-resources-tackbp-1.pdf}
  {{Overview of Linguistic Resources for the {TAC} {KBP} 2016 Evaluations:
  {M}ethodologies and Results}}.
\newblock In \emph{TAC}.

\bibitem[{Ellis et~al.(2014)Ellis, Getman, and Strassel}]{ellis2014overview}
Joe Ellis, Jeremy Getman, and Stephanie~M Strassel. 2014.
\newblock \href
  {https://www.ldc.upenn.edu/sites/www.ldc.upenn.edu/files/tackbp-2014-overview.pdf}
  {Overview of linguistic resources for the {TAC} {KBP} 2014 evaluations:
  {P}lanning, execution, and results}.
\newblock In \emph{TAC}.

\bibitem[{Espinosa et~al.(2019)Espinosa, Miwa, and
  Ananiadou}]{espinosa-etal-2019-search}
Kurt~Junshean Espinosa, Makoto Miwa, and Sophia Ananiadou. 2019.
\newblock \href {https://doi.org/10.18653/v1/D19-1381} {A search-based neural
  model for biomedical nested and overlapping event detection}.
\newblock In \emph{Proceedings of EMNLP-IJCNLP}, pages 3679--3686.

\bibitem[{Feng et~al.(2016)Feng, Huang, Tang, Ji, Qin, and
  Liu}]{feng-etal-2016-language}
Xiaocheng Feng, Lifu Huang, Duyu Tang, Heng Ji, Bing Qin, and Ting Liu. 2016.
\newblock \href {https://doi.org/10.18653/v1/P16-2011} {A language-independent
  neural network for event detection}.
\newblock In \emph{Proceedings of ACL}, pages 66--71.

\bibitem[{Ge et~al.(2016)Ge, Cui, Chang, Sui, and Zhou}]{ge-etal-2016-event}
Tao Ge, Lei Cui, Baobao Chang, Zhifang Sui, and Ming Zhou. 2016.
\newblock \href {https://aclanthology.org/C16-1309} {Event detection with burst
  information networks}.
\newblock In \emph{Proceedings of {COLING}}, pages 3276--3286.

\bibitem[{Getman et~al.(2017)Getman, Ellis, Song, Tracey, and
  Strassel}]{getman2017overview}
Jeremy Getman, Joe Ellis, Zhiyi Song, Jennifer Tracey, and Stephanie Strassel.
  2017.
\newblock \href
  {https://tac.nist.gov/publications/2017/additional.papers/TAC2017.KBP_resources_overview.proceedings.pdf}
  {Overview of linguistic resources for the tac kbp 2017 evaluations:
  {M}ethodologies and results.}
\newblock In \emph{TAC}.

\bibitem[{Ghaeini et~al.(2016)Ghaeini, Fern, Huang, and
  Tadepalli}]{ghaeini-etal-2016-event}
Reza Ghaeini, Xiaoli Fern, Liang Huang, and Prasad Tadepalli. 2016.
\newblock \href {https://doi.org/10.18653/v1/P16-2060} {Event nugget detection
  with forward-backward recurrent neural networks}.
\newblock In \emph{Proceedings of ACL}, pages 369--373.

\bibitem[{Glava{\v{s}} and {\v{S}}najder(2014)}]{glavavs2014event}
Goran Glava{\v{s}} and Jan {\v{S}}najder. 2014.
\newblock \href
  {https://www.sciencedirect.com/science/article/abs/pii/S0957417414001985}
  {Event graphs for information retrieval and multi-document summarization}.
\newblock \emph{Expert systems with applications}, 41(15):6904--6916.

\bibitem[{Gupta and Ji(2009)}]{gupta-ji:2009:Short}
Prashant Gupta and Heng Ji. 2009.
\newblock \href {http://www.aclweb.org/anthology/P/P09/P09-2093} {{Predicting
  Unknown Time Arguments based on Cross-Event Propagation}}.
\newblock In \emph{Proceedings of ACL-IJCNLP}, pages 369--372.

\bibitem[{Guzman-Nateras et~al.(2022)Guzman-Nateras, Nguyen, and
  Nguyen}]{guzman-nateras-etal-2022-cross}
Luis Guzman-Nateras, Minh~Van Nguyen, and Thien Nguyen. 2022.
\newblock \href {https://doi.org/10.18653/v1/2022.naacl-main.409}
  {Cross-lingual event detection via optimized adversarial training}.
\newblock In \emph{Proceedings of NAACL-HLT}, pages 5588--5599.

\bibitem[{Hogenboom et~al.(2016)Hogenboom, Frasincar, Kaymak, De~Jong, and
  Caron}]{hogenboom2016survey}
Frederik Hogenboom, Flavius Frasincar, Uzay Kaymak, Franciska De~Jong, and
  Emiel Caron. 2016.
\newblock \href
  {https://www.sciencedirect.com/science/article/abs/pii/S0167923616300173?via%3Dihub}
  {A survey of event extraction methods from text for decision support
  systems}.
\newblock \emph{Decision Support Systems}, 85:12--22.

\bibitem[{Hong et~al.(2011)Hong, Zhang, Ma, Yao, Zhou, and
  Zhu}]{hong-etal-2011-using}
Yu~Hong, Jianfeng Zhang, Bin Ma, Jianmin Yao, Guodong Zhou, and Qiaoming Zhu.
  2011.
\newblock \href {https://aclanthology.org/P11-1113} {Using cross-entity
  inference to improve event extraction}.
\newblock In \emph{Proceedings of ACL-HLT}, pages 1127--1136.

\bibitem[{Hsi et~al.(2016)Hsi, Yang, Carbonell, and
  Xu}]{hsi-etal-2016-leveraging}
Andrew Hsi, Yiming Yang, Jaime Carbonell, and Ruochen Xu. 2016.
\newblock \href {https://aclanthology.org/C16-1114} {Leveraging multilingual
  training for limited resource event extraction}.
\newblock In \emph{Proceedings of {COLING}}, pages 1201--1210.

\bibitem[{Hsu et~al.(2022)Hsu, Huang, Boschee, Miller, Natarajan, Chang, and
  Peng}]{hsu-etal-2022-degree}
I-Hung Hsu, Kuan-Hao Huang, Elizabeth Boschee, Scott Miller, Prem Natarajan,
  Kai-Wei Chang, and Nanyun Peng. 2022.
\newblock \href {https://doi.org/10.18653/v1/2022.naacl-main.138} {{DEGREE}:
  {A} data-efficient generation-based event extraction model}.
\newblock In \emph{Proceedings of NAACL-HLT}, pages 1890--1908.

\bibitem[{Huang et~al.(2022)Huang, Hsu, Natarajan, Chang, and
  Peng}]{huang-etal-2022-multilingual-generative}
Kuan-Hao Huang, I-Hung Hsu, Prem Natarajan, Kai-Wei Chang, and Nanyun Peng.
  2022.
\newblock \href {https://doi.org/10.18653/v1/2022.acl-long.317} {Multilingual
  generative language models for zero-shot cross-lingual event argument
  extraction}.
\newblock In \emph{Proceedings of ACL}, pages 4633--4646.

\bibitem[{Huang et~al.(2020{\natexlab{a}})Huang, Yang, and
  Peng}]{huang-etal-2020-biomedical}
Kung-Hsiang Huang, Mu~Yang, and Nanyun Peng. 2020{\natexlab{a}}.
\newblock \href {https://doi.org/10.18653/v1/2020.findings-emnlp.114}
  {Biomedical event extraction with hierarchical knowledge graphs}.
\newblock In \emph{Findings of EMNLP}, pages 1277--1285.

\bibitem[{Huang et~al.(2016)Huang, Cassidy, Feng, Ji, Voss, Han, and
  Sil}]{huang-etal-2016-liberal}
Lifu Huang, Taylor Cassidy, Xiaocheng Feng, Heng Ji, Clare~R. Voss, Jiawei Han,
  and Avirup Sil. 2016.
\newblock \href {https://doi.org/10.18653/v1/P16-1025} {{Liberal Event
  Extraction and Event Schema Induction}}.
\newblock In \emph{Proceedings of ACL}, pages 258--268.

\bibitem[{Huang and Ji(2020)}]{huang-ji-2020-semi}
Lifu Huang and Heng Ji. 2020.
\newblock \href {https://doi.org/10.18653/v1/2020.emnlp-main.53}
  {{Semi-supervised New Event Type Induction and Event Detection}}.
\newblock In \emph{Proceedings of EMNLP}, pages 718--724.

\bibitem[{Huang et~al.(2020{\natexlab{b}})Huang, Zhao, Takanobu, Tan, and
  Xiao}]{huang-etal-2020-joint}
Peixin Huang, Xiang Zhao, Ryuichi Takanobu, Zhen Tan, and Weidong Xiao.
  2020{\natexlab{b}}.
\newblock \href {https://doi.org/10.18653/v1/2020.coling-main.239} {Joint event
  extraction with hierarchical policy network}.
\newblock In \emph{Proceedings of {COLING}}, pages 2653--2664.

\bibitem[{Huang and Jia(2021)}]{huang-jia-2021-exploring-sentence}
Yusheng Huang and Weijia Jia. 2021.
\newblock \href {https://doi.org/10.18653/v1/2021.findings-emnlp.32} {Exploring
  sentence community for document-level event extraction}.
\newblock In \emph{Findings of EMNLP}, pages 340--351.

\bibitem[{Intxaurrondo et~al.(2015)Intxaurrondo, Agirre, Lopez~de Lacalle, and
  Surdeanu}]{intxaurrondo-etal-2015-diamonds}
Ander Intxaurrondo, Eneko Agirre, Oier Lopez~de Lacalle, and Mihai Surdeanu.
  2015.
\newblock \href {https://doi.org/10.3115/v1/N15-1066} {Diamonds in the rough:
  {E}vent extraction from imperfect microblog data}.
\newblock In \emph{Proceedings of NAACL-HLT}, pages 641--650.

\bibitem[{Jagannatha and Yu(2016)}]{jagannatha-yu-2016-bidirectional}
Abhyuday~N Jagannatha and Hong Yu. 2016.
\newblock \href {https://doi.org/10.18653/v1/N16-1056} {Bidirectional {RNN} for
  medical event detection in electronic health records}.
\newblock In \emph{Proceedings of NAACL-HLT}, pages 473--482.

\bibitem[{Ji and Grishman(2008)}]{ji2008refining}
Heng Ji and Ralph Grishman. 2008.
\newblock \href {http://aclweb.org/anthology/P08-1030} {Refining event
  extraction through cross-document inference}.
\newblock In \emph{Proceedings of ACL}, pages 254--262.

\bibitem[{Ji and Grishman(2011)}]{ji-grishman-2011-knowledge}
Heng Ji and Ralph Grishman. 2011.
\newblock \href {https://www.aclweb.org/anthology/P11-1115} {{Knowledge Base
  Population: {S}uccessful Approaches and Challenges}}.
\newblock In \emph{Proceedings of ACL}, pages 1148--1158.

\bibitem[{Judea and Strube(2016)}]{judea-strube-2016-incremental}
Alex Judea and Michael Strube. 2016.
\newblock \href {https://aclanthology.org/C16-1215} {Incremental global event
  extraction}.
\newblock In \emph{Proceedings of {COLING}}, pages 2279--2289.

\bibitem[{Lafferty et~al.(2001)Lafferty, McCallum, and Pereira}]{crf}
John~D. Lafferty, Andrew McCallum, and Fernando C.~N. Pereira. 2001.
\newblock \href {https://dl.acm.org/doi/abs/10.5555/645530.655813} {Conditional
  random fields: {P}robabilistic models for segmenting and labeling sequence
  data}.
\newblock In \emph{Proceedings of ICML}, pages 282--289.

\bibitem[{Lai et~al.(2021)Lai, Dernoncourt, and
  Nguyen}]{lai-etal-2021-learning}
Viet Lai, Franck Dernoncourt, and Thien~Huu Nguyen. 2021.
\newblock \href {https://doi.org/10.18653/v1/2021.emnlp-main.427} {Learning
  prototype representations across few-shot tasks for event detection}.
\newblock In \emph{Proceedings of EMNLP}, pages 5270--5277.

\bibitem[{Lai et~al.(2020)Lai, Nguyen, and Nguyen}]{lai-etal-2020-event}
Viet~Dac Lai, Tuan~Ngo Nguyen, and Thien~Huu Nguyen. 2020.
\newblock \href {https://doi.org/10.18653/v1/2020.emnlp-main.435} {{Event
  Detection: {G}ate Diversity and Syntactic Importance Scores for Graph
  Convolution Neural Networks}}.
\newblock In \emph{Proceedings of EMNLP}, pages 5405--5411.

\bibitem[{Lee et~al.(2015)Lee, Artzi, Choi, and
  Zettlemoyer}]{lee-etal-2015-event}
Kenton Lee, Yoav Artzi, Yejin Choi, and Luke Zettlemoyer. 2015.
\newblock \href {https://doi.org/10.18653/v1/D15-1189} {Event detection and
  factuality assessment with non-expert supervision}.
\newblock In \emph{Proceedings of EMNLP}, pages 1643--1648.

\bibitem[{Lewis et~al.(2020)Lewis, Liu, Goyal, Ghazvininejad, Mohamed, Levy,
  Stoyanov, and Zettlemoyer}]{lewis-etal-2020-bart}
Mike Lewis, Yinhan Liu, Naman Goyal, Marjan Ghazvininejad, Abdelrahman Mohamed,
  Omer Levy, Veselin Stoyanov, and Luke Zettlemoyer. 2020.
\newblock \href {https://doi.org/10.18653/v1/2020.acl-main.703} {{BART}:
  {D}enoising sequence-to-sequence pre-training for natural language
  generation, translation, and comprehension}.
\newblock In \emph{Proceedings of ACL}, pages 7871--7880.

\bibitem[{Li et~al.(2019)Li, Huang, Ji, and Han}]{li-etal-2019-biomedical}
Diya Li, Lifu Huang, Heng Ji, and Jiawei Han. 2019.
\newblock \href {https://doi.org/10.18653/v1/N19-1145} {Biomedical event
  extraction based on knowledge-driven tree-{LSTM}}.
\newblock In \emph{Proceedings of NAACL-HLT}, pages 1421--1430.

\bibitem[{Li et~al.(2020{\natexlab{a}})Li, Peng, Chen, Wang, Pan, Lyu, and
  Zhu}]{li-etal-2020-event}
Fayuan Li, Weihua Peng, Yuguang Chen, Quan Wang, Lu~Pan, Yajuan Lyu, and Yong
  Zhu. 2020{\natexlab{a}}.
\newblock \href {https://doi.org/10.18653/v1/2020.findings-emnlp.73} {Event
  extraction as multi-turn question answering}.
\newblock In \emph{Findings of EMNLP}, pages 829--838.

\bibitem[{Li et~al.(2022{\natexlab{a}})Li, Mo, Fan, Wang, Wang, Zhang, and
  Li}]{li-kipt-2022}
Haochen Li, Tong Mo, Hongcheng Fan, Jingkun Wang, Jiaxi Wang, Fuhao Zhang, and
  Weiping Li. 2022{\natexlab{a}}.
\newblock \href {https://aclanthology.org/2022.coling-1.169} {{KiPT}:
  {K}nowledge-injected prompt tuning for event detection}.
\newblock In \emph{Proceedings of {COLING}}, pages 1943--1952.

\bibitem[{Li et~al.(2013)Li, Ji, and Huang}]{li2013joint}
Qi~Li, Heng Ji, and Liang Huang. 2013.
\newblock \href {http://aclweb.org/anthology/P13-1008} {Joint event extraction
  via structured prediction with global features}.
\newblock In \emph{Proceedings of ACL}, pages 73--82.

\bibitem[{Li et~al.(2021{\natexlab{a}})Li, Zhao, Yang, and
  Su}]{li-treasure-2021}
Rui Li, Wenlin Zhao, Cheng Yang, and Sen Su. 2021{\natexlab{a}}.
\newblock \href {https://doi.org/10.18653/v1/2021.emnlp-main.206} {Treasures
  outside contexts: {I}mproving event detection via global statistics}.
\newblock In \emph{Proceedings of EMNLP}, pages 2625--2635.

\bibitem[{Li et~al.(2021{\natexlab{b}})Li, Ji, and Han}]{li-etal-2021-document}
Sha Li, Heng Ji, and Jiawei Han. 2021{\natexlab{b}}.
\newblock \href {https://doi.org/10.18653/v1/2021.naacl-main.69}
  {Document-level event argument extraction by conditional generation}.
\newblock In \emph{Proceedings of NAACL-HLT}, pages 894--908.

\bibitem[{Li et~al.(2020{\natexlab{b}})Li, Li, Pan, Chen, Peng, Wang, Lyu, and
  Zhu}]{duee}
Xinyu Li, Fayuan Li, Lu~Pan, Yuguang Chen, Weihua Peng, Quan Wang, Yajuan Lyu,
  and Yong Zhu. 2020{\natexlab{b}}.
\newblock \href {https://doi.org/10.1007/978-3-030-60457-8\_44} {Duee: {A}
  large-scale dataset for chinese event extraction in real-world scenarios}.
\newblock In \emph{Proceedings of NLPCC}, volume 12431 of \emph{Lecture Notes
  in Computer Science}, pages 534--545.

\bibitem[{Li et~al.(2022{\natexlab{b}})Li, Hong, Wang, He, Yao, and
  Zhou}]{li-etal-2022-unregulated}
Zhongqiu Li, Yu~Hong, Jie Wang, Shiming He, Jianmin Yao, and Guodong Zhou.
  2022{\natexlab{b}}.
\newblock \href {https://aclanthology.org/2022.coling-1.232} {Unregulated
  {C}hinese-to-{E}nglish data expansion does {NOT} work for neural event
  detection}.
\newblock In \emph{Proceedings of {COLING}}, pages 2633--2638.

\bibitem[{Lin et~al.(2019)Lin, Lu, Han, and Sun}]{lin-etal-2019-cost}
Hongyu Lin, Yaojie Lu, Xianpei Han, and Le~Sun. 2019.
\newblock \href {https://doi.org/10.18653/v1/P19-1521} {Cost-sensitive
  regularization for label confusion-aware event detection}.
\newblock In \emph{Proceedings of ACL}, pages 5278--5283.

\bibitem[{Lin et~al.(2020)Lin, Ji, Huang, and Wu}]{lin-etal-2020-joint}
Ying Lin, Heng Ji, Fei Huang, and Lingfei Wu. 2020.
\newblock \href {https://doi.org/10.18653/v1/2020.acl-main.713} {A joint neural
  model for information extraction with global features}.
\newblock In \emph{Proceedings of ACL}, pages 7999--8009.

\bibitem[{Liu et~al.(2021{\natexlab{a}})Liu, Xu, and
  Liu}]{liu-etal-2021-self-attention-graph}
Anan Liu, Ning Xu, and Haozhe Liu. 2021{\natexlab{a}}.
\newblock \href {https://doi.org/10.18653/v1/2021.findings-emnlp.28}
  {Self-attention graph residual convolutional networks for event detection
  with dependency relations}.
\newblock In \emph{Findings of EMNLP}, pages 302--311.

\bibitem[{Liu et~al.(2020{\natexlab{a}})Liu, Chen, Liu, Bi, and
  Liu}]{liu-etal-2020-event}
Jian Liu, Yubo Chen, Kang Liu, Wei Bi, and Xiaojiang Liu. 2020{\natexlab{a}}.
\newblock \href {https://doi.org/10.18653/v1/2020.emnlp-main.128} {{Event
  Extraction as Machine Reading Comprehension}}.
\newblock In \emph{Proceedings of EMNLP}, pages 1641--1651.

\bibitem[{Liu et~al.(2020{\natexlab{b}})Liu, Chen, Liu, Jia, and
  Sheng}]{liu-etal-2020-context}
Jian Liu, Yubo Chen, Kang Liu, Yantao Jia, and Zhicheng Sheng.
  2020{\natexlab{b}}.
\newblock \href {https://doi.org/10.18653/v1/2020.findings-emnlp.229} {How does
  context matter? {O}n the robustness of event detection with context-selective
  mask generalization}.
\newblock In \emph{Findings of EMNLP}, pages 2523--2532.

\bibitem[{Liu et~al.(2019{\natexlab{a}})Liu, Chen, Liu, and
  Zhao}]{liu-etal-2019-neural}
Jian Liu, Yubo Chen, Kang Liu, and Jun Zhao. 2019{\natexlab{a}}.
\newblock \href {https://doi.org/10.18653/v1/D19-1068} {Neural cross-lingual
  event detection with minimal parallel resources}.
\newblock In \emph{Proceedings of EMNLP-IJCNLP}, pages 738--748.

\bibitem[{Liu et~al.(2021{\natexlab{b}})Liu, Chen, and
  Xu}]{liu-etal-2021-machine}
Jian Liu, Yufeng Chen, and Jinan Xu. 2021{\natexlab{b}}.
\newblock \href {https://doi.org/10.18653/v1/2021.emnlp-main.214} {Machine
  reading comprehension as data augmentation: {A} case study on implicit event
  argument extraction}.
\newblock In \emph{Proceedings of EMNLP}, pages 2716--2725.

\bibitem[{Liu et~al.(2022{\natexlab{a}})Liu, Chen, and
  Xu}]{liu-etal-2022-saliency}
Jian Liu, Yufeng Chen, and Jinan Xu. 2022{\natexlab{a}}.
\newblock \href {https://doi.org/10.18653/v1/2022.acl-long.313} {Saliency as
  evidence: {E}vent detection with trigger saliency attribution}.
\newblock In \emph{Proceedings of ACL}, pages 4573--4585.

\bibitem[{Liu et~al.(2022{\natexlab{b}})Liu, Chang, and
  Huang}]{liu-etal-2022-incremental}
Minqian Liu, Shiyu Chang, and Lifu Huang. 2022{\natexlab{b}}.
\newblock \href {https://aclanthology.org/2022.coling-1.189} {Incremental
  prompting: {E}pisodic memory prompt for lifelong event detection}.
\newblock In \emph{Proceedings of {COLING}}, pages 2157--2165.

\bibitem[{Liu et~al.(2018{\natexlab{a}})Liu, Cheng, Yu, and
  Cheng}]{liu-etal-2018-exploiting-contextual}
Shaobo Liu, Rui Cheng, Xiaoming Yu, and Xueqi Cheng. 2018{\natexlab{a}}.
\newblock \href {https://doi.org/10.18653/v1/D18-1127} {Exploiting contextual
  information via dynamic memory network for event detection}.
\newblock In \emph{Proceedings of EMNLP}, pages 1030--1035.

\bibitem[{Liu et~al.(2016)Liu, Chen, He, Liu, and
  Zhao}]{liu-etal-2016-leveraging}
Shulin Liu, Yubo Chen, Shizhu He, Kang Liu, and Jun Zhao. 2016.
\newblock \href {https://doi.org/10.18653/v1/P16-1201} {Leveraging {F}rame{N}et
  to improve automatic event detection}.
\newblock In \emph{Proceedings of ACL}, pages 2134--2143.

\bibitem[{Liu et~al.(2017)Liu, Chen, Liu, and Zhao}]{liu-etal-2017-exploiting}
Shulin Liu, Yubo Chen, Kang Liu, and Jun Zhao. 2017.
\newblock \href {https://doi.org/10.18653/v1/P17-1164} {{Exploiting Argument
  Information to Improve Event Detection via Supervised Attention Mechanisms}}.
\newblock In \emph{Proceedings of ACL}, pages 1789--1798.

\bibitem[{Liu et~al.(2019{\natexlab{b}})Liu, Li, Zhang, Yang, and
  Zhou}]{liu-etal-2019-event}
Shulin Liu, Yang Li, Feng Zhang, Tao Yang, and Xinpeng Zhou.
  2019{\natexlab{b}}.
\newblock \href {https://doi.org/10.18653/v1/N19-1080} {Event detection without
  triggers}.
\newblock In \emph{Proceedings of NAACL-HLT}, pages 735--744.

\bibitem[{Liu et~al.(2022{\natexlab{c}})Liu, Huang, Shi, and
  Wang}]{liu-etal-2022-dynamic}
Xiao Liu, Heyan Huang, Ge~Shi, and Bo~Wang. 2022{\natexlab{c}}.
\newblock \href {https://doi.org/10.18653/v1/2022.acl-long.358} {Dynamic
  prefix-tuning for generative template-based event extraction}.
\newblock In \emph{Proceedings of ACL}, pages 5216--5228.

\bibitem[{Liu et~al.(2018{\natexlab{b}})Liu, Luo, and
  Huang}]{liu-etal-2018-jointly}
Xiao Liu, Zhunchen Luo, and Heyan Huang. 2018{\natexlab{b}}.
\newblock \href {https://doi.org/10.18653/v1/D18-1156} {Jointly multiple events
  extraction via attention-based graph information aggregation}.
\newblock In \emph{Proceedings of EMNLP}, pages 1247--1256.

\bibitem[{Lou et~al.(2021)Lou, Liao, Deng, Zhang, and
  Chen}]{lou-etal-2021-mlbinet}
Dongfang Lou, Zhilin Liao, Shumin Deng, Ningyu Zhang, and Huajun Chen. 2021.
\newblock \href {https://doi.org/10.18653/v1/2021.acl-long.373} {{MLB}i{N}et:
  {A} cross-sentence collective event detection network}.
\newblock In \emph{Proceedings of ACL-IJCNLP}, pages 4829--4839.

\bibitem[{Lu and Nguyen(2018)}]{lu-nguyen-2018-similar}
Weiyi Lu and Thien~Huu Nguyen. 2018.
\newblock \href {https://doi.org/10.18653/v1/D18-1517} {Similar but not the
  same: {W}ord sense disambiguation improves event detection via neural
  representation matching}.
\newblock In \emph{Proceedings of EMNLP}, pages 4822--4828.

\bibitem[{Lu et~al.(2019)Lu, Lin, Han, and Sun}]{lu-etal-2019-distilling}
Yaojie Lu, Hongyu Lin, Xianpei Han, and Le~Sun. 2019.
\newblock \href {https://doi.org/10.18653/v1/P19-1429} {Distilling
  discrimination and generalization knowledge for event detection via
  delta-representation learning}.
\newblock In \emph{Proceedings of ACL}, pages 4366--4376.

\bibitem[{Lu et~al.(2021)Lu, Lin, Xu, Han, Tang, Li, Sun, Liao, and
  Chen}]{lu-etal-2021-text2event}
Yaojie Lu, Hongyu Lin, Jin Xu, Xianpei Han, Jialong Tang, Annan Li, Le~Sun,
  Meng Liao, and Shaoyi Chen. 2021.
\newblock \href {https://doi.org/10.18653/v1/2021.acl-long.217}
  {{T}ext2{E}vent: {C}ontrollable sequence-to-structure generation for
  end-to-end event extraction}.
\newblock In \emph{Proceedings of ACL-IJCNLP}, pages 2795--2806.

\bibitem[{Lyu et~al.(2021)Lyu, Zhang, Sulem, and Roth}]{lyu-etal-2021-zero}
Qing Lyu, Hongming Zhang, Elior Sulem, and Dan Roth. 2021.
\newblock \href {https://doi.org/10.18653/v1/2021.acl-short.42} {Zero-shot
  event extraction via transfer learning: {C}hallenges and insights}.
\newblock In \emph{Proceedings of ACL-IJCNLP}, pages 322--332.

\bibitem[{Ma et~al.(2020)Ma, Wang, Anubhai, Ballesteros, and
  Al-Onaizan}]{ma-etal-2020-resource}
Jie Ma, Shuai Wang, Rishita Anubhai, Miguel Ballesteros, and Yaser Al-Onaizan.
  2020.
\newblock \href {https://doi.org/10.18653/v1/2020.findings-emnlp.318}
  {Resource-enhanced neural model for event argument extraction}.
\newblock In \emph{Findings of EMNLP}, pages 3554--3559.

\bibitem[{Ma et~al.(2022)Ma, Wang, Cao, Li, Chen, Wang, and Shao}]{paie}
Yubo Ma, Zehao Wang, Yixin Cao, Mukai Li, Meiqi Chen, Kun Wang, and Jing Shao.
  2022.
\newblock \href {https://doi.org/10.18653/v1/2022.acl-long.466} {Prompt for
  extraction? {PAIE:} {P}rompting argument interaction for event argument
  extraction}.
\newblock In \emph{Proceedings of ACL}, pages 6759--6774.

\bibitem[{Man Duc~Trong et~al.(2020)Man Duc~Trong, Trong~Le, Pouran Ben~Veyseh,
  Nguyen, and Nguyen}]{man-duc-trong-etal-2020-introducing}
Hieu Man Duc~Trong, Duc Trong~Le, Amir Pouran Ben~Veyseh, Thuat Nguyen, and
  Thien~Huu Nguyen. 2020.
\newblock \href {https://doi.org/10.18653/v1/2020.emnlp-main.433} {Introducing
  a new dataset for event detection in cybersecurity texts}.
\newblock In \emph{Proceedings of EMNLP}, pages 5381--5390.

\bibitem[{Mi et~al.(2022)Mi, Hu, and Li}]{mi-etal-2022-event}
Jiaxin Mi, Po~Hu, and Peng Li. 2022.
\newblock \href {https://aclanthology.org/2022.coling-1.172} {Event detection
  with dual relational graph attention networks}.
\newblock In \emph{Proceedings of {COLING}}, pages 1979--1989.

\bibitem[{Naik and Rose(2020)}]{naik-rose-2020-towards}
Aakanksha Naik and Carolyn Rose. 2020.
\newblock \href {https://doi.org/10.18653/v1/2020.acl-main.681} {Towards open
  domain event trigger identification using adversarial domain adaptation}.
\newblock In \emph{Proceedings of ACL}, pages 7618--7624.

\bibitem[{Ngo~Trung et~al.(2021)Ngo~Trung, Phung, and
  Nguyen}]{ngo-trung-etal-2021-unsupervised}
Nghia Ngo~Trung, Duy Phung, and Thien~Huu Nguyen. 2021.
\newblock \href {https://doi.org/10.18653/v1/2021.findings-acl.351}
  {Unsupervised domain adaptation for event detection using domain-specific
  adapters}.
\newblock In \emph{Findings of ACL-IJCNLP}, pages 4015--4025.

\bibitem[{Nguyen et~al.(2022)Nguyen, Min, Dernoncourt, and
  Nguyen}]{nguyen-etal-2022-joint}
Minh~Van Nguyen, Bonan Min, Franck Dernoncourt, and Thien Nguyen. 2022.
\newblock \href {https://doi.org/10.18653/v1/2022.naacl-main.324} {Joint
  extraction of entities, relations, and events via modeling inter-instance and
  inter-label dependencies}.
\newblock In \emph{Proceedings of NAACL-HLT}, pages 4363--4374.

\bibitem[{Nguyen et~al.(2021)Nguyen, Nguyen, Min, and
  Nguyen}]{nguyen-etal-2021-crosslingual}
Minh~Van Nguyen, Tuan~Ngo Nguyen, Bonan Min, and Thien~Huu Nguyen. 2021.
\newblock \href {https://doi.org/10.18653/v1/2021.emnlp-main.440} {Crosslingual
  transfer learning for relation and event extraction via word category and
  class alignments}.
\newblock In \emph{Proceedings of EMNLP}, pages 5414--5426.

\bibitem[{Nguyen and Grishman(2018)}]{nguyen2018graph}
Thien Nguyen and Ralph Grishman. 2018.
\newblock \href
  {https://www.aaai.org/ocs/index.php/AAAI/AAAI18/paper/view/16329} {Graph
  convolutional networks with argument-aware pooling for event detection}.
\newblock In \emph{Proceedings of AAAI}, pages 5900--5907.

\bibitem[{Nguyen et~al.(2016)Nguyen, Cho, and
  Grishman}]{nguyen-etal-2016-joint-event}
Thien~Huu Nguyen, Kyunghyun Cho, and Ralph Grishman. 2016.
\newblock \href {https://doi.org/10.18653/v1/N16-1034} {Joint event extraction
  via recurrent neural networks}.
\newblock In \emph{Proceedings of NAACL-HLT}, pages 300--309.

\bibitem[{Nguyen and Grishman(2015)}]{nguyen-grishman-2015-event}
Thien~Huu Nguyen and Ralph Grishman. 2015.
\newblock \href {https://doi.org/10.3115/v1/P15-2060} {{{Event Detection and
  Domain Adaptation with Convolutional Neural Networks}}}.
\newblock In \emph{Proceedings of ACL}, pages 365--371.

\bibitem[{Nguyen and Grishman(2016)}]{nguyen-grishman-2016-modeling}
Thien~Huu Nguyen and Ralph Grishman. 2016.
\newblock \href {https://doi.org/10.18653/v1/D16-1085} {Modeling skip-grams for
  event detection with convolutional neural networks}.
\newblock In \emph{Proceedings of EMNLP}, pages 886--891.

\bibitem[{Orr et~al.(2018)Orr, Tadepalli, and Fern}]{orr-etal-2018-event}
Walker Orr, Prasad Tadepalli, and Xiaoli Fern. 2018.
\newblock \href {https://doi.org/10.18653/v1/D18-1122} {Event detection with
  neural networks: {A} rigorous empirical evaluation}.
\newblock In \emph{Proceedings of EMNLP}, pages 999--1004.

\bibitem[{Peng et~al.(2016)Peng, Song, and Roth}]{peng-etal-2016-event}
Haoruo Peng, Yangqiu Song, and Dan Roth. 2016.
\newblock \href {https://doi.org/10.18653/v1/D16-1038} {Event detection and
  co-reference with minimal supervision}.
\newblock In \emph{Proceedings of EMNLP}, pages 392--402.

\bibitem[{Pouran Ben~Veyseh et~al.(2021{\natexlab{a}})Pouran Ben~Veyseh, Lai,
  Dernoncourt, and Nguyen}]{pouran-ben-veyseh-etal-2021-unleash}
Amir Pouran Ben~Veyseh, Viet Lai, Franck Dernoncourt, and Thien~Huu Nguyen.
  2021{\natexlab{a}}.
\newblock \href {https://doi.org/10.18653/v1/2021.acl-long.490} {Unleash
  {GPT}-2 power for event detection}.
\newblock In \emph{Proceedings of ACL-IJCNLP}, pages 6271--6282.

\bibitem[{Pouran Ben~Veyseh et~al.(2022)Pouran Ben~Veyseh, Nguyen, Dernoncourt,
  Min, and Nguyen}]{pouran-ben-veyseh-etal-2022-document}
Amir Pouran Ben~Veyseh, Minh~Van Nguyen, Franck Dernoncourt, Bonan Min, and
  Thien Nguyen. 2022.
\newblock \href {https://doi.org/10.18653/v1/2022.findings-acl.130}
  {Document-level event argument extraction via optimal transport}.
\newblock In \emph{Findings of ACL}, pages 1648--1658.

\bibitem[{Pouran Ben~Veyseh et~al.(2021{\natexlab{b}})Pouran Ben~Veyseh,
  Nguyen, Ngo~Trung, Min, and Nguyen}]{pouran-ben-veyseh-etal-2021-modeling}
Amir Pouran Ben~Veyseh, Minh~Van Nguyen, Nghia Ngo~Trung, Bonan Min, and
  Thien~Huu Nguyen. 2021{\natexlab{b}}.
\newblock \href {https://doi.org/10.18653/v1/2021.emnlp-main.439} {Modeling
  document-level context for event detection via important context selection}.
\newblock In \emph{Proceedings of EMNLP}, pages 5403--5413.

\bibitem[{Pouran Ben~Veyseh et~al.(2020)Pouran Ben~Veyseh, Nguyen, and
  Nguyen}]{pouran-ben-veyseh-etal-2020-graph}
Amir Pouran Ben~Veyseh, Tuan~Ngo Nguyen, and Thien~Huu Nguyen. 2020.
\newblock \href {https://doi.org/10.18653/v1/2020.findings-emnlp.326} {{Graph
  Transformer Networks with Syntactic and Semantic Structures for Event
  Argument Extraction}}.
\newblock In \emph{Findings of EMNLP}, pages 3651--3661.

\bibitem[{Raffel et~al.(2020)Raffel, Shazeer, Roberts, Lee, Narang, Matena,
  Zhou, Li, and Liu}]{2020t5}
Colin Raffel, Noam Shazeer, Adam Roberts, Katherine Lee, Sharan Narang, Michael
  Matena, Yanqi Zhou, Wei Li, and Peter~J. Liu. 2020.
\newblock \href {http://jmlr.org/papers/v21/20-074.html} {Exploring the limits
  of transfer learning with a unified text-to-text transformer}.
\newblock \emph{Journal of Machine Learning Research}, 21(140):1--67.

\bibitem[{Ramponi et~al.(2020)Ramponi, van~der Goot, Lombardo, and
  Plank}]{ramponi-etal-2020-biomedical}
Alan Ramponi, Rob van~der Goot, Rosario Lombardo, and Barbara Plank. 2020.
\newblock \href {https://doi.org/10.18653/v1/2020.emnlp-main.431} {Biomedical
  event extraction as sequence labeling}.
\newblock In \emph{Proceedings of EMNLP}, pages 5357--5367.

\bibitem[{Ramshaw and Marcus(1995)}]{ramshaw1999text}
Lance Ramshaw and Mitch Marcus. 1995.
\newblock \href {https://aclanthology.org/W95-0107} {Text chunking using
  transformation-based learning}.
\newblock In \emph{Third Workshop on Very Large Corpora}.

\bibitem[{Ren et~al.(2022)Ren, Cao, Fang, Guo, Lin, Ma, and
  Liu}]{ren-etal-2022-clio}
Yubing Ren, Yanan Cao, Fang Fang, Ping Guo, Zheng Lin, Wei Ma, and Yi~Liu.
  2022.
\newblock \href {https://aclanthology.org/2022.coling-1.221} {{CLIO}:
  {R}ole-interactive multi-event head attention network for document-level
  event extraction}.
\newblock In \emph{Proceedings of {COLING}}, pages 2504--2514.

\bibitem[{Sainz et~al.(2022)Sainz, Gonzalez-Dios, Lopez~de Lacalle, Min, and
  Agirre}]{sainz-etal-2022-textual}
Oscar Sainz, Itziar Gonzalez-Dios, Oier Lopez~de Lacalle, Bonan Min, and Eneko
  Agirre. 2022.
\newblock \href {https://doi.org/10.18653/v1/2022.findings-naacl.187} {Textual
  entailment for event argument extraction: {Z}ero- and few-shot with
  multi-source learning}.
\newblock In \emph{Findings of NAACL-HLT}, pages 2439--2455.

\bibitem[{Sha et~al.(2016)Sha, Liu, Lin, Li, Chang, and
  Sui}]{sha-etal-2016-rbpb}
Lei Sha, Jing Liu, Chin-Yew Lin, Sujian Li, Baobao Chang, and Zhifang Sui.
  2016.
\newblock \href {https://doi.org/10.18653/v1/P16-1116} {{RBPB}:
  {R}egularization-based pattern balancing method for event extraction}.
\newblock In \emph{Proceedings of ACL}, pages 1224--1234.

\bibitem[{Shen et~al.(2020)Shen, Qi, Li, Bi, and
  Wang}]{shen-etal-2020-hierarchical}
Shirong Shen, Guilin Qi, Zhen Li, Sheng Bi, and Lusheng Wang. 2020.
\newblock \href {https://doi.org/10.18653/v1/2020.coling-main.9} {Hierarchical
  {C}hinese legal event extraction via pedal attention mechanism}.
\newblock In \emph{Proceedings of {COLING}}, pages 100--113.

\bibitem[{Shen et~al.(2021)Shen, Wu, Qi, Li, Haffari, and
  Bi}]{shen-etal-2021-adaptive}
Shirong Shen, Tongtong Wu, Guilin Qi, Yuan-Fang Li, Gholamreza Haffari, and
  Sheng Bi. 2021.
\newblock \href {https://doi.org/10.18653/v1/2021.findings-acl.214} {Adaptive
  knowledge-enhanced {B}ayesian meta-learning for few-shot event detection}.
\newblock In \emph{Findings of ACL-IJCNLP}, pages 2417--2429.

\bibitem[{Sheng et~al.(2021)Sheng, Guo, Yu, Li, Hei, Wang, Liu, and
  Xu}]{sheng-etal-2021-casee}
Jiawei Sheng, Shu Guo, Bowen Yu, Qian Li, Yiming Hei, Lihong Wang, Tingwen Liu,
  and Hongbo Xu. 2021.
\newblock \href {https://doi.org/10.18653/v1/2021.findings-acl.14} {{C}as{EE}:
  {A} joint learning framework with cascade decoding for overlapping event
  extraction}.
\newblock In \emph{Findings of ACL-IJCNLP}, pages 164--174.

\bibitem[{Sims et~al.(2019)Sims, Park, and Bamman}]{sims-etal-2019-literary}
Matthew Sims, Jong~Ho Park, and David Bamman. 2019.
\newblock \href {https://doi.org/10.18653/v1/P19-1353} {Literary event
  detection}.
\newblock In \emph{Proceedings of ACL}, pages 3623--3634.

\bibitem[{Song et~al.(2015)Song, Bies, Strassel, Riese, Mott, Ellis, Wright,
  Kulick, Ryant, and Ma}]{song2015light}
Zhiyi Song, Ann Bies, Stephanie Strassel, Tom Riese, Justin Mott, Joe Ellis,
  Jonathan Wright, Seth Kulick, Neville Ryant, and Xiaoyi Ma. 2015.
\newblock \href {https://aclanthology.org/W15-0812} {From light to rich ere:
  {A}nnotation of entities, relations, and events}.
\newblock In \emph{Proceedings of the 3rd Workshop on EVENTS: Definition,
  Detection, Coreference, and Representation}, pages 89--98.

\bibitem[{Subburathinam et~al.(2019)Subburathinam, Lu, Ji, May, Chang, Sil, and
  Voss}]{subburathinam-etal-2019-cross}
Ananya Subburathinam, Di~Lu, Heng Ji, Jonathan May, Shih-Fu Chang, Avirup Sil,
  and Clare Voss. 2019.
\newblock \href {https://doi.org/10.18653/v1/D19-1030} {Cross-lingual structure
  transfer for relation and event extraction}.
\newblock In \emph{Proceedings of EMNLP-IJCNLP}, pages 313--325.

\bibitem[{Tong et~al.(2020)Tong, Xu, Wang, Cao, Hou, Li, and
  Xie}]{tong-etal-2020-improving}
Meihan Tong, Bin Xu, Shuai Wang, Yixin Cao, Lei Hou, Juanzi Li, and Jun Xie.
  2020.
\newblock \href {https://doi.org/10.18653/v1/2020.acl-main.522} {Improving
  event detection via open-domain trigger knowledge}.
\newblock In \emph{Proceedings of ACL}, pages 5887--5897.

\bibitem[{Wadden et~al.(2019)Wadden, Wennberg, Luan, and
  Hajishirzi}]{wadden-etal-2019-entity}
David Wadden, Ulme Wennberg, Yi~Luan, and Hannaneh Hajishirzi. 2019.
\newblock \href {https://doi.org/10.18653/v1/D19-1585} {{Entity, Relation, and
  Event Extraction with Contextualized Span Representations}}.
\newblock In \emph{Proceedings of EMNLP-IJCNLP}, pages 5784--5789.

\bibitem[{Walker et~al.(2006)Walker, Strassel, Medero, and
  Maeda}]{walker2006ace}
Christopher Walker, Stephanie Strassel, Julie Medero, and Kazuaki Maeda. 2006.
\newblock \href {https://catalog.ldc.upenn.edu/LDC2006T06} {{ACE} 2005
  multilingual training corpus}.
\newblock \emph{Linguistic Data Consortium}, 57.

\bibitem[{Wang et~al.(2022)Wang, Yu, Chang, Sun, and
  Huang}]{wang-etal-2022-query}
Sijia Wang, Mo~Yu, Shiyu Chang, Lichao Sun, and Lifu Huang. 2022.
\newblock \href {https://doi.org/10.18653/v1/2022.findings-acl.16} {Query and
  extract: {R}efining event extraction as type-oriented binary decoding}.
\newblock In \emph{Findings of ACL}, pages 169--182.

\bibitem[{Wang et~al.(2019{\natexlab{a}})Wang, Han, Liu, Sun, and
  Li}]{wang-etal-2019-adversarial-training}
Xiaozhi Wang, Xu~Han, Zhiyuan Liu, Maosong Sun, and Peng Li.
  2019{\natexlab{a}}.
\newblock \href {https://doi.org/10.18653/v1/N19-1105} {{Adversarial Training
  for Weakly Supervised Event Detection}}.
\newblock In \emph{Proceedings of NAACL-HLT}, pages 998--1008.

\bibitem[{Wang et~al.(2020)Wang, Wang, Han, Jiang, Han, Liu, Li, Li, Lin, and
  Zhou}]{wang-etal-2020-maven}
Xiaozhi Wang, Ziqi Wang, Xu~Han, Wangyi Jiang, Rong Han, Zhiyuan Liu, Juanzi
  Li, Peng Li, Yankai Lin, and Jie Zhou. 2020.
\newblock \href {https://doi.org/10.18653/v1/2020.emnlp-main.129} {{MAVEN}: {A}
  {M}assive {G}eneral {D}omain {E}vent {D}etection {D}ataset}.
\newblock In \emph{Proceedings of EMNLP}, pages 1652--1671.

\bibitem[{Wang et~al.(2019{\natexlab{b}})Wang, Wang, Han, Liu, Li, Li, Sun,
  Zhou, and Ren}]{wang-etal-2019-hmeae}
Xiaozhi Wang, Ziqi Wang, Xu~Han, Zhiyuan Liu, Juanzi Li, Peng Li, Maosong Sun,
  Jie Zhou, and Xiang Ren. 2019{\natexlab{b}}.
\newblock \href {https://doi.org/10.18653/v1/D19-1584} {{HMEAE}: {H}ierarchical
  {Modular Event Argument Extraction}}.
\newblock In \emph{Proceedings of EMNLP-IJCNLP}, pages 5777--5783.

\bibitem[{Wang et~al.(2021)Wang, Wang, Han, Lin, Hou, Liu, Li, Li, and
  Zhou}]{wang-etal-2021-cleve}
Ziqi Wang, Xiaozhi Wang, Xu~Han, Yankai Lin, Lei Hou, Zhiyuan Liu, Peng Li,
  Juanzi Li, and Jie Zhou. 2021.
\newblock \href {https://doi.org/10.18653/v1/2021.acl-long.491} {{CLEVE}:
  {C}ontrastive {P}re-training for {E}vent {E}xtraction}.
\newblock In \emph{Proceedings of ACL-IJCNLP}, pages 6283--6297.

\bibitem[{Wei et~al.(2021)Wei, Sun, Zhang, Zhang, Zhi, and
  Jin}]{wei-etal-2021-trigger}
Kaiwen Wei, Xian Sun, Zequn Zhang, Jingyuan Zhang, Guo Zhi, and Li~Jin. 2021.
\newblock \href {https://doi.org/10.18653/v1/2021.acl-long.360} {Trigger is not
  sufficient: {E}xploiting frame-aware knowledge for implicit event argument
  extraction}.
\newblock In \emph{Proceedings of ACL-IJCNLP}, pages 4672--4682.

\bibitem[{Wei et~al.(2017)Wei, Korostil, Nothman, and
  Hachey}]{wei-etal-2017-english}
Sam Wei, Igor Korostil, Joel Nothman, and Ben Hachey. 2017.
\newblock \href {https://doi.org/10.18653/v1/P17-2046} {{E}nglish event
  detection with translated language features}.
\newblock In \emph{Proceedings of ACL}, pages 293--298.

\bibitem[{Wei et~al.(2022)Wei, Liu, Lv, Xi, Yan, Ye, Mo, Yang, and
  Wan}]{wei-etal-2022-desed}
Yinyi Wei, Shuaipeng Liu, Jianwei Lv, Xiangyu Xi, Hailei Yan, Wei Ye, Tong Mo,
  Fan Yang, and Guanglu Wan. 2022.
\newblock \href {https://aclanthology.org/2022.coling-1.219} {{DESED}:
  {D}ialogue-based explanation for sentence-level event detection}.
\newblock In \emph{Proceedings of {COLING}}, pages 2483--2493.

\bibitem[{Wurzer et~al.(2015)Wurzer, Lavrenko, and
  Osborne}]{wurzer-etal-2015-twitter}
Dominik Wurzer, Victor Lavrenko, and Miles Osborne. 2015.
\newblock \href {https://doi.org/10.18653/v1/D15-1310} {{T}witter-scale new
  event detection via k-term hashing}.
\newblock In \emph{Proceedings of EMNLP}, pages 2584--2589.

\bibitem[{Xi et~al.(2021)Xi, Ye, Zhang, Wang, Jiang, and
  Wu}]{xi-etal-2021-capturing}
Xiangyu Xi, Wei Ye, Shikun Zhang, Quanxiu Wang, Huixing Jiang, and Wei Wu.
  2021.
\newblock \href {https://doi.org/10.18653/v1/2021.acl-long.18} {Capturing event
  argument interaction via a bi-directional entity-level recurrent decoder}.
\newblock In \emph{Proceedings of ACL-IJCNLP}, pages 210--219.

\bibitem[{Xie et~al.(2021)Xie, Sun, Zhou, Qu, and Dai}]{xie-etal-2021-event}
Jianye Xie, Haotong Sun, Junsheng Zhou, Weiguang Qu, and Xinyu Dai. 2021.
\newblock \href {https://doi.org/10.18653/v1/2021.findings-acl.142} {Event
  detection as graph parsing}.
\newblock In \emph{Findings of ACL-IJCNLP}, pages 1630--1640.

\bibitem[{Xu et~al.(2021)Xu, Liu, Li, and Chang}]{xu-etal-2021-document}
Runxin Xu, Tianyu Liu, Lei Li, and Baobao Chang. 2021.
\newblock \href {https://doi.org/10.18653/v1/2021.acl-long.274} {Document-level
  event extraction via heterogeneous graph-based interaction model with a
  tracker}.
\newblock In \emph{Proceedings of ACL-IJCNLP}, pages 3533--3546.

\bibitem[{Xu et~al.(2022)Xu, Wang, Liu, Zeng, Chang, and
  Sui}]{xu-etal-2022-two}
Runxin Xu, Peiyi Wang, Tianyu Liu, Shuang Zeng, Baobao Chang, and Zhifang Sui.
  2022.
\newblock \href {https://doi.org/10.18653/v1/2022.naacl-main.370} {A two-stream
  {AMR}-enhanced model for document-level event argument extraction}.
\newblock In \emph{Proceedings of NAACL-HLT}, pages 5025--5036.

\bibitem[{Yagcioglu et~al.(2019)Yagcioglu, Seyfioglu, Citamak, Bardak,
  Guldamlasioglu, Yuksel, and Tatli}]{yagcioglu-etal-2019-detecting}
Semih Yagcioglu, Mehmet~Saygin Seyfioglu, Begum Citamak, Batuhan Bardak, Seren
  Guldamlasioglu, Azmi Yuksel, and Emin~Islam Tatli. 2019.
\newblock \href {https://doi.org/10.18653/v1/N19-1138} {Detecting cybersecurity
  events from noisy short text}.
\newblock In \emph{Proceedings of NAACL-HLT}, pages 1366--1372.

\bibitem[{Yan et~al.(2019)Yan, Jin, Meng, Guo, and Cheng}]{yan-etal-2019-event}
Haoran Yan, Xiaolong Jin, Xiangbin Meng, Jiafeng Guo, and Xueqi Cheng. 2019.
\newblock \href {https://doi.org/10.18653/v1/D19-1582} {{Event Detection with
  Multi-Order Graph Convolution and Aggregated Attention}}.
\newblock In \emph{Proceedings of EMNLP-IJCNLP}, pages 5766--5770.

\bibitem[{Yang and Mitchell(2016)}]{yang-mitchell-2016-joint}
Bishan Yang and Tom~M. Mitchell. 2016.
\newblock \href {https://doi.org/10.18653/v1/N16-1033} {Joint extraction of
  events and entities within a document context}.
\newblock In \emph{Proceedings of NAACL-HLT}, pages 289--299.

\bibitem[{Yang et~al.(2021)Yang, Sui, Chen, Liu, Zhao, and
  Wang}]{yang-etal-2021-document}
Hang Yang, Dianbo Sui, Yubo Chen, Kang Liu, Jun Zhao, and Taifeng Wang. 2021.
\newblock \href {https://doi.org/10.18653/v1/2021.acl-long.492} {Document-level
  event extraction via parallel prediction networks}.
\newblock In \emph{Proceedings of ACL-IJCNLP}, pages 6298--6308.

\bibitem[{Yang et~al.(2019)Yang, Feng, Qiao, Kan, and
  Li}]{yang-etal-2019-exploring}
Sen Yang, Dawei Feng, Linbo Qiao, Zhigang Kan, and Dongsheng Li. 2019.
\newblock \href {https://doi.org/10.18653/v1/P19-1522} {Exploring pre-trained
  language models for event extraction and generation}.
\newblock In \emph{Proceedings of ACL}, pages 5284--5294.

\bibitem[{Yao et~al.(2022)Yao, Xiao, Wang, Liu, Hou, Tu, Li, Liu, Shen, and
  Sun}]{leven}
Feng Yao, Chaojun Xiao, Xiaozhi Wang, Zhiyuan Liu, Lei Hou, Cunchao Tu, Juanzi
  Li, Yun Liu, Weixing Shen, and Maosong Sun. 2022.
\newblock \href {https://doi.org/10.18653/v1/2022.findings-acl.17} {{LEVEN:}
  {A} large-scale chinese legal event detection dataset}.
\newblock In \emph{Findings of ACL}, pages 183--201.

\bibitem[{Yu et~al.(2021)Yu, Ji, and Natarajan}]{yu-etal-2021-lifelong}
Pengfei Yu, Heng Ji, and Prem Natarajan. 2021.
\newblock \href {https://doi.org/10.18653/v1/2021.emnlp-main.428} {Lifelong
  event detection with knowledge transfer}.
\newblock In \emph{Proceedings of EMNLP}, pages 5278--5290.

\bibitem[{Zeng et~al.(2022)Zeng, Zhan, and Ji}]{zeng-etal-2022-ea2e}
Qi~Zeng, Qiusi Zhan, and Heng Ji. 2022.
\newblock \href {https://doi.org/10.18653/v1/2022.findings-naacl.202}
  {{EA}$^2${E}: {I}mproving consistency with event awareness for document-level
  argument extraction}.
\newblock In \emph{Findings of NAACL}, pages 2649--2655.

\bibitem[{Zhang et~al.(2020{\natexlab{a}})Zhang, Liu, Pan, Song, and
  Leung}]{zhang2020aser}
Hongming Zhang, Xin Liu, Haojie Pan, Yangqiu Song, and Cane Wing-Ki Leung.
  2020{\natexlab{a}}.
\newblock \href {https://dl.acm.org/doi/abs/10.1145/3366423.3380107} {{ASER}:
  {A} large-scale eventuality knowledge graph}.
\newblock In \emph{Proceedings of WWW}, pages 201--211.

\bibitem[{Zhang et~al.(2021)Zhang, Wang, and Roth}]{zhang-etal-2021-zero}
Hongming Zhang, Haoyu Wang, and Dan Roth. 2021.
\newblock \href {https://doi.org/10.18653/v1/2021.findings-acl.114}
  {{Z}ero-shot {L}abel-aware {E}vent {T}rigger and {A}rgument
  {C}lassification}.
\newblock In \emph{Findings of ACL-IJCNLP}, pages 1331--1340.

\bibitem[{Zhang et~al.(2022)Zhang, Ji, Ji, and Wang}]{zhang-etal-2022-zero}
Senhui Zhang, Tao Ji, Wendi Ji, and Xiaoling Wang. 2022.
\newblock \href {https://doi.org/10.18653/v1/2022.findings-naacl.196}
  {Zero-shot event detection based on ordered contrastive learning and
  prompt-based prediction}.
\newblock In \emph{Findings of NAACL-HLT}, pages 2572--2580.

\bibitem[{Zhang et~al.(2020{\natexlab{b}})Zhang, Kong, Liu, Ma, and
  Hovy}]{zhang-etal-2020-two}
Zhisong Zhang, Xiang Kong, Zhengzhong Liu, Xuezhe Ma, and Eduard Hovy.
  2020{\natexlab{b}}.
\newblock \href {https://doi.org/10.18653/v1/2020.acl-main.667} {A two-step
  approach for implicit event argument detection}.
\newblock In \emph{Proceedings of ACL}, pages 7479--7485.

\bibitem[{Zhang and Ji(2021)}]{zhang-ji-2021-abstract}
Zixuan Zhang and Heng Ji. 2021.
\newblock \href {https://doi.org/10.18653/v1/2021.naacl-main.4} {{A}bstract
  {M}eaning {R}epresentation guided graph encoding and decoding for joint
  information extraction}.
\newblock In \emph{Proceedings of NAACL-HLT}, pages 39--49.

\bibitem[{Zheng et~al.(2019)Zheng, Cao, Xu, and
  Bian}]{zheng-etal-2019-doc2edag}
Shun Zheng, Wei Cao, Wei Xu, and Jiang Bian. 2019.
\newblock \href {https://doi.org/10.18653/v1/D19-1032} {{D}oc2{EDAG}: {A}n
  end-to-end document-level framework for {C}hinese financial event
  extraction}.
\newblock In \emph{Proceedings of EMNLP-IJCNLP}, pages 337--346.

\bibitem[{Zhou and Mao(2022)}]{zhou-mao-2022-document}
Hanzhang Zhou and Kezhi Mao. 2022.
\newblock \href {https://doi.org/10.18653/v1/2022.naacl-main.222}
  {Document-level event argument extraction by leveraging redundant information
  and closed boundary loss}.
\newblock In \emph{Proceedings of NAACL-HLT}, pages 3041--3052.

\bibitem[{Zhou et~al.(2022)Zhou, Zhang, Chen, Zhang, He, and
  Huang}]{zhou-etal-2022-multi}
Jie Zhou, Qi~Zhang, Qin Chen, Qi~Zhang, Liang He, and Xuanjing Huang. 2022.
\newblock \href {https://aclanthology.org/2022.coling-1.173} {A multi-format
  transfer learning model for event argument extraction via variational
  information bottleneck}.
\newblock In \emph{Proceedings of COLING}, pages 1990--2000.

\bibitem[{Zhou et~al.(2021)Zhou, Chen, Zhao, Wu, Xu, and Li}]{zhou2021role}
Yang Zhou, Yubo Chen, Jun Zhao, Yin Wu, Jiexin Xu, and Jinlong Li. 2021.
\newblock \href {https://ojs.aaai.org/index.php/AAAI/article/view/17720} {What
  the role is vs. what plays the role: {S}emi-supervised event argument
  extraction via dual question answering}.
\newblock In \emph{Proceedings of AAAI}, volume~35, pages 14638--14646.

\end{thebibliography}
\bibliographystyle{acl_natbib}

\appendix
\clearpage
\section*{Appendices}

\section{Experimental Details}
\label{sec:experimental_details}
The section introduces the experimental details in the paper, including the data preprocessing details (\cref{sec:data_preprocessing_details}), the reproduction details (\cref{sec:reproduce_details}), and the training details (\cref{sec:training_details}).

\subsection{Data Preprocessing Details}
\label{sec:data_preprocessing_details}
The section introduces the details of the three data preprocessing scripts for ACE 2005: \acedygie, \aceoneie, and \acefull. 
\paragraph{ACE-DYGIE}
We adopt the released official codes\footnote{\url{https://github.com/dwadden/dygiepp}} provided by~\citet{wadden-etal-2019-entity} as the \acedygie preprocessing script. Specifically, we adopt the widely-used ``default-settings'' in the codes to preprocess ACE 2005. \acedygie uses spaCy\footnote{\url{https://spacy.io/}} for sentence segmentation and tokenization. The version of spaCy is $2.0.18$, and the used spaCy model is \texttt{en\_core\_web\_sm}.

\paragraph{ACE-OneIE}
We adopt the released official codes\footnote{\url{https://blender.cs.illinois.edu/software/oneie/}} provided by~\citet{lin-etal-2020-joint} as the \aceoneie preprocessing script. \aceoneie uses NLTK\footnote{\url{https://www.nltk.org/}} for sentence segmentation and tokenization , and the version of NLTK is $3.5$.

\paragraph{ACE-Full}
We adopt the released official codes\footnote{\url{https://github.com/thunlp/HMEAE}} provided by~\citet{wang-etal-2019-hmeae} as the \acefull preprocessing script. \acefull uses the Stanford CoreNLP\footnote{\url{https://stanfordnlp.github.io/CoreNLP/}} toolkit for sentence segmentation and tokenization, and the version of CoreNLP is $4.4.0$.

\subsection{Reproduction Details}
\label{sec:reproduce_details}
In this section, we introduce the reproduction details of all the reproduced models and provide some explanations for the results' differences between our reproduction and the originally reported results. All the reproduction experiments adopt their original evaluation settings, respectively. The number of parameters for each reproduced model is shown in Table~\ref{tab:num_param}. 

\begin{table}
    \centering
    \small
    \begin{tabular}{lr}
    \toprule
    \textbf{Model} & \textbf{\#Paramter} \\
    \midrule
    DMCNN  & $2$M \\
    DMBERT & $110$M \\
    CLEVE & $354$M \\
    BiLSTM+CRF & $37$M \\
    BERT+CRF & $110$M \\
    EEQA & $110$M \\
    PAIE & $406$M \\
    Text2Event & $770$M \\
    \bottomrule
    \end{tabular}
    \caption{Number of parameters for each reproduced model.}
    \label{tab:num_param}
\end{table}

\paragraph{DMCNN}
Our DMCNN implementation is mainly based on the codes\footnote{\url{https://github.com/THU-KEG/MAVEN-dataset}\label{maven}} provided by~\citet{wang-etal-2020-maven}. The reproduced ED F1 score ($67.2$) is similar to the reported result ($69.1$) in the original paper~\citep{chen2015event} on the ACE 2005 dataset. However, there is a gap between our reproduced and the originally reported EAE F1 scores ($43.2$ vs. $53.5$). A possible reason is that \citet{chen2015event} adopts a different EAE evaluation setting: Only the argument annotations of the predicted triggers are included in the metric calculation, while the argument annotations of the false negative trigger predictions are discarded. This setting is also adopted in some other early works like DMBERT~\cite{wang-etal-2019-hmeae}, and we call it ``legacy setting''. Compared to the common evaluation setting now, which includes all the argument annotations, the recall scores under the legacy setting are typically higher. When re-evaluating our reproduced DMCNN under the legacy setting, the EAE F1 score ($53.9$) is consistent with the originally reported result ($53.5$).

\paragraph{DMBERT}
Our DMBERT implementation is mainly based on the codes\textsuperscript{\ref{maven}} provided by~\cite{wang-etal-2020-maven}. The reproduced ED F1 score ($74.5$) is consistent with the originally reported result ($74.3$) on the ACE 2005 dataset. However, similar to the DMCNN case introduced in the last paragraph, the reproduced EAE F1 score ($54.8$) is lower than the originally reported result ($57.2$ in \citet{wang-etal-2019-hmeae}) due to the ``legacy setting''. When re-evaluating the reproduced DMBERT under the legacy setting, the EAE F1 score is $60.6$.  

\paragraph{CLEVE}
We download the pre-trained CLEVE checkpoint\footnote{\url{https://github.com/THU-KEG/CLEVE}} and finetune it on ACE 2005. The reproduced F1 scores of ED ($78.3$) and EAE ($61.0$) are basically consistent with the originally reported ED ($79.8$) and EAE ($61.1$) results. 

\paragraph{BiLSTM+CRF}
We implement BiLSTM+CRF based on the codes\textsuperscript{\ref{maven}} provided by~\citet{wang-etal-2020-maven}. The reproduced ED F1 score ($75.5$) is similar to the reported result ($75.4$) in the original paper~\citep{wang-etal-2020-maven} on ACE 2005. As there is no work using BiLSTM+CRF to perform EAE, we adopt all the settings used in ED and evaluate the EAE performance of BiLSTM+CRF.

\paragraph{BERT+CRF}
We implement BERT+CRF based on the codes\textsuperscript{\ref{maven}} provided by~\citet{wang-etal-2020-maven}. The reproduced ED F1 score ($72.1$) is similar to the reported result ($74.1$) in the original paper~\citep{wang-etal-2020-maven} on ACE 2005. As there is no work using BERT+CRF to perform EAE, we implement its EAE model following all the ED settings.

\paragraph{EEQA}
We implement EEQA~\citep{du-cardie-2020-event} based on the released official codes\footnote{\url{https://github.com/xinyadu/eeqa}}. When directly running the released code, we get the F1 score of $69.0$ for ED and $47.3$ for EAE, which are consistent with our finally reproduced ED ($69.5$) and EAE ($47.4$) results. However, there is still a gap between the reproduced and the originally reported results, which is also mentioned in several GitHub issues\footnote{\url{https://github.com/xinyadu/eeqa/issues/11}, \url{https://github.com/xinyadu/eeqa/issues/5}}.

\paragraph{PAIE}
We implement PAIE~\citep{paie} based on the released official codes\footnote{\url{https://github.com/mayubo2333/PAIE}} and evaluate it in different evaluation settings. The reproduced EAE F1 score ($71.8$) is basically consistent with that reported in the original paper ($72.7$).

\paragraph{Text2Event}
We adopt the released official codes\footnote{\url{https://github.com/luyaojie/Text2Event}} to re-evaluate Text2Event~\citep{lu-etal-2021-text2event} in different settings. There are minor differences between the reproduced F1 results and the originally reported results (ED: $69.5$ vs. $71.9$, EAE: $50.8$ vs. $53.8$). We think the differences come from randomness. When only using the same random seed reported by the authors, the reproduction results are nearly the same as the original results.

\subsection{Training Details}
\label{sec:training_details}
We run three random trials for all the experiments using three different seeds (seed=$0$, seed=$1$, seed=$2$). The final reported results are the average results over the three random trials. All hyper-parameters are the same as those used in the original papers.
The experiments of CLEVE, PAIE, and Text2Event are run on Nvidia A100 GPUs, which consume about $600$ GPU hours. The other experiments are run on Nvidia GeForce RTX 3090 GPUs, which consume about $100$ GPU hours.
\section{Additional Experimental Results}
\label{sec:app_add_exp}
The section shows additional experimental results on different preprocessed ACE 2005 datasets. 

\paragraph{Output Space Discrepancy}
\Cref{tab:appendix_output_discrepancy} shows the metrics' differences with and without output standardization on the \acedygie and \acefull preprocessed datasets. We can observe that all evaluation metrics change obviously, which is consistent with the observations in \cref{sec:output_space}.

\begin{table*}
    \centering
    \small
    \begin{tabular}{l|ccc|ccc|ccc|ccc}
    \toprule 
    & \multicolumn{6}{c|}{\textbf{\acedygie}} & 
    \multicolumn{6}{c}{\textbf{\acefull}} \\
    \cmidrule{2-13}
    & \multicolumn{3}{c|}{\textbf{ED}} & \multicolumn{3}{c|}{\textbf{EAE}} & \multicolumn{3}{c|}{\textbf{ED}} &
    \multicolumn{3}{c}{\textbf{EAE}} \\
    \cmidrule{1-13}
    \textbf{Metric} & $\Delta$\textbf{P} & $\Delta$\textbf{R} & $\Delta$\textbf{F1} 
    & $\Delta$\textbf{P} & $\Delta$\textbf{R} & $\Delta$\textbf{F1} & $\Delta$\textbf{P} & $\Delta$\textbf{R} & $\Delta$\textbf{F1} 
    & $\Delta$\textbf{P} & $\Delta$\textbf{R} & $\Delta$\textbf{F1} \\ 
    \midrule
    BiLSTM+CRF & $+0.4$& $+0.0$& $+0.2$& $+6.7$& $+0.2$& $+3.7$ & $+1.9$& $-0.2$& $+1.0$& $+15.7$& $+0.1$& $+7.4$ \\
    BERT+CRF & $+0.6$& $-0.2$& $+0.2$& $+5.2$& $-0.1$& $+2.9$ & $+2.5$& $-0.2$& $+1.3$& $+14.1$& $-0.4$& $+6.1$ \\
    EEQA & $+0.0$& $+0.0$& $+0.0$& $-0.7$& $-1.2$& $-1.0$ & $-0.3$& $+1.3$& $+0.4$& $+19.3$& $-15.3$& $-6.9$ \\
    PAIE & \texttt{N/A} & \texttt{N/A} & \texttt{N/A} & $+4.4$& $-0.5$& $+1.9$ & \texttt{N/A} & \texttt{N/A} & \texttt{N/A} & $+21.9$& $-1.6$& $+8.4$ \\
    Text2Event & $+0.3$& $+0.0$& $+0.1$& $+0.5$& $-2.5$& $-0.9$ & $+2.0$& $+0.0$& $+1.0$& $+6.2$& $-3.7$& $+0.1$ \\
    \bottomrule
    \end{tabular}
    \caption{The precision, recall, and F1 (\%) differences between evaluation with and without our output standardization. The results are evaluated on the \acedygie and \acefull preprocessed datasets. Output standardization aligns the output spaces of the other paradigms into that of the CLS paradigm, and hence we do not include the CLS-paradigm models here, whose results are unchanged.}
    \label{tab:appendix_output_discrepancy}
\end{table*}

\paragraph{Absence of Pipeline Evaluation}
\Cref{tab:appendix_pipeline} shows the results using gold trigger evaluation and pipeline evaluation on the \acedygie and \acefull preprocessed datasets. We can observe that the phenomena are consistent with those in 
\cref{sec:absence_pipeline}.

\begin{table*}[!t]
    \centering
    \small
    \begin{tabular}{l|ccc|ccc}
    \toprule
    & \multicolumn{3}{c|}{\textbf{\acedygie}} &
    \multicolumn{3}{c}{\textbf{\acefull}} \\
    \cmidrule{1-7}
    \multirow{2}{*}{\textbf{Metric}} & \multirow{2}{*}{\textbf{ED F1}} & \textbf{Gold Tri.} & \textbf{Pipeline} & \multirow{2}{*}{\textbf{ED F1}} & \textbf{Gold Tri.} & \textbf{Pipeline} \\
    & & \textbf{EAE F1} & \textbf{EAE F1} & & \textbf{EAE F1}& \textbf{EAE F1} \\
    \midrule
    DMCNN & $62.5$& $50.1$& $34.0$ & $67.2$& $61.8$& $43.2$ \\
    DMBERT  & $68.3$& $67.3$& $48.1$ & $74.5$& $73.1$& $54.8$ \\
    CLEVE  & $72.9$& $71.4$& $54.8$ & $78.3$& $76.2$& $61.0$ \\
    BiLSTM+CRF  & $72.0$& $45.2$& $36.2$ & $76.5$& $46.2$& $36.9$ \\
    BERT+CRF  & $68.1$& $64.1$& $47.8$& $73.4$& $64.5$& $48.6$ \\
    EEQA  & $69.5$& $63.5$& $46.4$& $73.6$& $46.1$& $36.4$ \\
    PAIE  & $72.9$ & $73.8$ & $56.5$& $78.3$ & $65.0$& $52.7$ \\
    \bottomrule
    \end{tabular}
    \caption{EAE F1 scores (\%) of gold trigger evaluation and pipeline evaluation on \acedygie and \acefull. We also report corresponding ED F1 scores to show trigger quality. PAIE adopts the triggers predicted by CLEVE. The joint model Text2Event is excluded since its trigger input cannot be controlled.}
    \label{tab:appendix_pipeline}
\end{table*}

\paragraph{Consistent Evaluation Framework}
\Cref{tab:appendix_consistent_eval} shows the results using our consistent evaluation on \acedygie, \aceoneie, and \acefull. We can observe that the phenomena on \acedygie and \aceoneie are consistent with those in \cref{sec:experimental_results}.

\begin{table*}
    \centering
    \small
    \begin{tabular}{l|ccc|ccc}
    \toprule
    & \multicolumn{3}{c|}{\textbf{ED}} & \multicolumn{3}{c}{\textbf{EAE}} \\
    \midrule
    \textbf{Metric} & \textbf{P} & \textbf{R} & \textbf{F1} & \textbf{P} & \textbf{R} & \textbf{F1} \\ 
    \midrule
    \multicolumn{7}{c}{\textbf{\acedygie}} \\
    \midrule
    DMCNN & $58.6\pm2.28$& $67.0\pm0.88$& $62.5\pm1.08$& $38.6\pm1.58$& $30.4\pm0.99$& $34.0\pm1.20$ \\
    DMBERT & $66.4\pm0.69$& $70.2\pm0.73$& $68.3\pm0.43$& $45.6\pm1.46$& $51.0\pm0.87$& $48.1\pm0.91$ \\
    CLEVE & $70.7\pm0.87$& $75.3\pm0.82$& $72.9\pm0.53$& $52.2\pm1.47$& $57.6\pm1.40$& $54.8\pm1.26$ \\
    BiLSTM+CRF & $68.5\pm1.27$& $75.8\pm2.28$& $72.0\pm0.99$& $36.4\pm1.21$& $36.1\pm0.37$& $36.2\pm0.55$ \\
    BERT+CRF & $64.0\pm1.94$& $72.8\pm1.57$& $68.1\pm1.01$& $46.3\pm1.35$& $49.5\pm2.04$& $47.8\pm0.70$ \\
    EEQA & $65.3\pm3.46$& $74.5\pm1.22$& $69.5\pm1.41$& $49.0\pm3.88$& $44.3\pm1.30$& $46.4\pm1.06$ \\
    PAIE & \texttt{N/A} & \texttt{N/A} & \texttt{N/A} & $56.5\pm0.49$& $56.5\pm1.28$& $56.5\pm0.87$ \\
    Text2Event & $67.2\pm0.82$& $72.4\pm0.62$& $69.7\pm0.72$& $48.5\pm2.60$& $51.6\pm1.04$& $50.0\pm0.89$ \\
    \midrule
    \multicolumn{7}{c}{\textbf{\aceoneie}} \\
    \midrule
    DMCNN & $61.5\pm2.66$& $64.5\pm2.86$& $62.8\pm0.40$& $36.7\pm2.48$& $34.1\pm1.88$& $35.2\pm0.22$ \\
    DMBERT & $64.4\pm2.89$& $75.4\pm3.21$& $69.4\pm1.36$& $41.5\pm1.84$& $54.7\pm1.42$& $47.2\pm0.99$ \\
    CLEVE & $72.3\pm1.86$& $78.0\pm0.91$& $75.0\pm0.81$& $52.1\pm1.99$& $57.6\pm0.47$& $54.7\pm1.31$ \\
    BiLSTM+CRF & $73.0\pm1.55$& $71.8\pm0.11$& $72.4\pm0.82$& $37.0\pm2.33$& $33.1\pm1.01$& $34.9\pm1.56$ \\
    BERT+CRF & $69.6\pm4.08$& $69.2\pm4.23$& $69.2\pm1.18$& $48.9\pm3.25$& $45.5\pm2.75$& $47.1\pm1.04$ \\
    EEQA & $66.7\pm1.73$& $71.8\pm2.51$& $69.1\pm0.28$& $50.1\pm1.73$& $41.0\pm1.92$& $45.0\pm0.70$ \\
    PAIE & \texttt{N/A} & \texttt{N/A} & \texttt{N/A} & $56.1\pm0.30$& $57.4\pm0.55$& $56.7\pm0.29$ \\
    Text2Event & $71.4\pm1.44$& $74.1\pm1.77$& $72.7\pm0.20$& $51.5\pm1.46$& $51.6\pm0.65$& $51.6\pm0.99$ \\
    \midrule
    \multicolumn{7}{c}{\textbf{\acefull}} \\
    \midrule
    DMCNN & $65.0\pm3.33$& $69.7\pm0.62$& $67.2\pm1.53$& $45.3\pm4.79$& $41.6\pm1.93$& $43.2\pm1.79$ \\
    DMBERT & $72.1\pm0.80$& $77.1\pm1.53$& $74.5\pm0.85$& $50.5\pm1.53$& $60.0\pm1.82$& $54.8\pm1.67$ \\
    CLEVE & $76.4\pm2.49$& $80.4\pm1.54$& $78.3\pm2.03$& $56.9\pm2.86$& $65.9\pm2.06$& $61.0\pm2.44$ \\
    BiLSTM+CRF & $74.2\pm1.62$& $78.9\pm0.45$& $76.5\pm1.02$& $42.8\pm1.20$& $32.4\pm0.23$& $36.9\pm0.60$ \\
    BERT+CRF & $72.4\pm2.34$& $74.5\pm1.23$& $73.4\pm1.29$& $55.6\pm1.51$& $43.2\pm1.31$& $48.6\pm0.96$ \\
    EEQA & $70.5\pm2.93$& $77.3\pm3.28$& $73.6\pm0.38$& $65.8\pm2.98$& $25.5\pm4.68$& $36.4\pm4.49$ \\
    PAIE & \texttt{N/A} & \texttt{N/A} & \texttt{N/A} & $61.4\pm1.70$& $46.2\pm0.64$& $52.7\pm0.77$ \\
    Text2Event & $76.1\pm0.25$& $74.5\pm1.28$& $75.2\pm0.68$& $59.6\pm0.96$& $43.0\pm1.49$& $50.0\pm1.07$ \\
    \bottomrule
    \end{tabular}
    \caption{Experimental results (\%) under our consistent evaluation on \acedygie, \aceoneie, and \acefull. We report averages and standard deviations over three runs. All the results are under pipeline evaluation. }
    \label{tab:appendix_consistent_eval}
\end{table*}
\section{Papers for Meta-Analysis}
\label{sec:review}
The complete list of papers surveyed in our meta-analysis is shown in \cref{tab:paper_list}.

\begin{table*}[!t]
    \centering
    \small
    \begin{adjustbox}{max width=1\linewidth}
    {
      \begin{tabular}{l}
        \toprule 
        ACL \\
        \midrule
        \citet{chen2015event}, \citet{bronstein-etal-2015-seed}, 
        \citet{nguyen-grishman-2015-event}\\
        \citet{sha-etal-2016-rbpb}, \citet{huang-etal-2016-liberal}
        \citet{ghaeini-etal-2016-event}, \citet{feng-etal-2016-language}, \citet{liu-etal-2016-leveraging}, \\
        \citet{wei-etal-2017-english}, \citet{liu-etal-2017-exploiting},
        \citet{chen-etal-2017-automatically},
        \\
        \citet{chan-etal-2019-rapid}, \citet{yang-etal-2019-exploring},
        \citet{sims-etal-2019-literary}, \citet{lu-etal-2019-distilling}, \citet{lin-etal-2019-cost}, \\ 
        \citet{lin-etal-2020-joint}, \citet{naik-rose-2020-towards}, \citet{tong-etal-2020-improving}, \\
        \citet{du-cardie-2020-document}, \citet{zhang-etal-2020-two}, \\
        \citet{zhang-etal-2021-zero}, \citet{lyu-etal-2021-zero}, \citet{ngo-trung-etal-2021-unsupervised}, \citet{pouran-ben-veyseh-etal-2021-unleash}, \\
        \citet{lu-etal-2021-text2event}, 
        \citet{deng-etal-2021-ontoed}, 
        \citet{lou-etal-2021-mlbinet}, \citet{cong-etal-2021-shot},\\
        \citet{xie-etal-2021-event}, 
        \citet{wang-etal-2021-cleve}, \citet{sheng-etal-2021-casee}, 
        \citet{shen-etal-2021-adaptive}, \\
        \citet{xi-etal-2021-capturing}, \citet{wei-etal-2021-trigger},
        \citet{yang-etal-2021-document}, \citet{xu-etal-2021-document}, \\
        \citet{liu-etal-2022-saliency}, \citet{wang-etal-2022-query}, \citet{liu-etal-2022-dynamic}, \citet{paie}, \\
        \citet{huang-etal-2022-multilingual-generative}, \citet{du-etal-2022-dynamic},
        \citet{pouran-ben-veyseh-etal-2022-document}\\
        
        \midrule 
        EMNLP \\
        \midrule
        \citet{wurzer-etal-2015-twitter}, \citet{lee-etal-2015-event}, \\
        \citet{nguyen-grishman-2016-modeling}, \citet{peng-etal-2016-event}, \\
        \citet{lu-nguyen-2018-similar}, \citet{chen-etal-2018-collective}, \citet{liu-etal-2018-exploiting-contextual}, \citet{orr-etal-2018-event}, \citet{liu-etal-2018-jointly}, \\
        \citet{liu-etal-2019-neural}, \citet{ding-etal-2019-event}, \citet{yan-etal-2019-event}, \citet{wang-etal-2019-hmeae},\\
        \citet{espinosa-etal-2019-search}, \citet{wadden-etal-2019-entity},
        \citet{zheng-etal-2019-doc2edag}, \citet{subburathinam-etal-2019-cross},\\
        \citet{du-cardie-2020-event}, \citet{huang-ji-2020-semi}, \citet{man-duc-trong-etal-2020-introducing}, \\
        \citet{cao-etal-2020-incremental}, \citet{liu-etal-2020-context}, \citet{li-etal-2020-event}, \citet{liu-etal-2020-event}, \citet{lai-etal-2020-event}, \citet{cui-etal-2020-edge}, \\
        \citet{huang-etal-2020-biomedical}, \citet{ramponi-etal-2020-biomedical}, \citet{ma-etal-2020-resource}, \citet{pouran-ben-veyseh-etal-2020-graph}, \\
        \citet{li-treasure-2021}, \citet{liu-etal-2021-self-attention-graph}, 
        \citet{pouran-ben-veyseh-etal-2021-modeling},\citet{yu-etal-2021-lifelong},\\
        \citet{lai-etal-2021-learning}, \citet{chen-etal-2021-honey},\citet{nguyen-etal-2021-crosslingual},
        \citet{liu-etal-2021-machine}, \citet{huang-jia-2021-exploring-sentence}\\
        \midrule
        NAACL \\
        \midrule
        \citet{intxaurrondo-etal-2015-diamonds},
        \\
        \citet{jagannatha-yu-2016-bidirectional}, \citet{yang-mitchell-2016-joint}, 
        \citet{nguyen-etal-2016-joint-event}, \\
        \citet{bekoulis-etal-2019-sub}, \citet{liu-etal-2019-event}, 
        \citet{yagcioglu-etal-2019-detecting}, \citet{li-etal-2019-biomedical},
        \citet{wang-etal-2019-adversarial-training}, \\
        \citet{zhang-ji-2021-abstract}, \citet{li-etal-2021-document}, \\
        \citet{zhang-etal-2022-zero}, \citet{nguyen-etal-2022-joint}, \citet{hsu-etal-2022-degree}, 
        \citet{guzman-nateras-etal-2022-cross}, \\
        \citet{sainz-etal-2022-textual}, \citet{zeng-etal-2022-ea2e},
        \citet{zhou-mao-2022-document}, \citet{xu-etal-2022-two}\\
        \midrule 
        COLING \\
        \midrule
        \citet{ge-etal-2016-event}, \citet{judea-strube-2016-incremental}, \citet{hsi-etal-2016-leveraging}, \\
        \citet{araki-mitamura-2018-open}, \\
        \citet{huang-etal-2020-joint},
        \citet{shen-etal-2020-hierarchical}, \\
        \citet{li-etal-2022-unregulated}, 
        \citet{ren-etal-2022-clio}, \citet{wei-etal-2022-desed},
        \citet{liu-etal-2022-incremental}, \\
        \citet{mi-etal-2022-event},
        \citet{cao-etal-2022-oneee}, \citet{li-kipt-2022}, \citet{zhou-etal-2022-multi} \\

        \bottomrule    
      \end{tabular}
    }
    \end{adjustbox}
    \caption{The complete list of papers for meta-analysis, categorized by venues and sorted by publication years.}
    \label{tab:paper_list}
\end{table*}%
\section{Authors' Contribution}
Hao Peng, Feng Yao, and Kaisheng Zeng conducted the empirical experiments. Feng Yao conducted the meta-analyses. Xiaozhi Wang, Hao Peng, and Feng Yao wrote the paper. Xiaozhi Wang designed the project. Lei Hou, Juanzi Li, Zhiyuan Liu, and Weixing Shen advised the project. All authors participated in the discussion.



\end{document}